%% file: main.tex
\documentclass[10pt,twocolumn,letterpaper]{article}

\usepackage{cvpr}              


\usepackage{url}            
\usepackage{booktabs}       
\usepackage{multirow}       
\usepackage{amsfonts}       
\usepackage{nicefrac}       
\usepackage{microtype}      
\usepackage{xcolor}         
\usepackage{lipsum}
\usepackage{amsmath}
\usepackage{amssymb}
\usepackage{graphicx}       
\usepackage{wrapfig}        
\usepackage{algorithm}
\usepackage{algorithmic}
\usepackage{makecell}
\usepackage[table]{xcolor}

\definecolor{cvprblue}{rgb}{0.21,0.49,0.74}
\usepackage[pagebackref,breaklinks,colorlinks,allcolors=cvprblue]{hyperref}

\def\confName{CVPR}
\def\confYear{2027}

\title{Pixel as the Common Language: Unified Multimodal Embedding via \\ Visual Rendering}

\author{
Xu Yuan, \ \ 
Pangjing Wu, \ \ 
Wenqi Fan\thanks{Corresponding Authors: Wenqi Fan and Qing Li}, \ \ 
Qing Li\footnotemark[1] \\ [0.1cm]
\small{The Hong Kong Polytechnic University, HK SAR} \\
\small{xuyuan127@gmail.com, pang-jing.wu@connect.polyu.hk, wenqifan03@gmail.com, qing-prof.li@polyu.edu.hk}\\
}

\begin{document}
\maketitle

\input{sections/0_abstract}
\input{sections/1_intro}
\input{sections/2_related_work}
\input{sections/3_method}
\input{sections/4_experiment}
\input{sections/5_conclusion}

{
    \small
    \bibliographystyle{ieeenat_fullname}
    \bibliography{main}
}

\clearpage
\appendix
\input{sections/appendix}

\end{document}

%% file: sections/0_abstract.tex
\begin{abstract}
Multimodal embedding models map heterogeneous inputs into a shared representation space, yet text-only, image-only, and composed inputs are still presented in distinct formats and follow different routes through the model. This fragmentation prevents heterogeneous tasks from jointly optimizing the same end-to-end encoding process, limiting supervision transfer across input configurations. To address this issue, we propose \textbf{unification before encoding} and instantiate this paradigm in \textbf{RenderEmbed}. The framework uses Visual Substrate Rendering to convert heterogeneous inputs into a common visual substrate for encoding with a VLM. From the resulting visual tokens, Progressive Hierarchical Anchor Encoding produces compact retrieval representations by interleaving learnable anchors and fusing their states across layers, retaining broad context and fine-grained cues. Interventional Rendering Learning trains these representations to remain sensitive to content changes and stable across rendering variations through complementary semantic and rendering interventions. Extensive experiments on MMEB and ViDoRe v2, spanning diverse retrieval tasks, show that RenderEmbed outperforms prior state-of-the-art models at the same scale and exhibits strong out-of-domain generalization. Compared with MetaEmbed, RenderEmbed improves MMEB Overall Precision@1 by 1.7 points, with a larger 2.1-point gain on out-of-domain tasks. Transfer and scaling analyses further show that a shared visual pathway enables more effective learning from heterogeneous supervision. Our code will be released at \url{https://github.com/xandery-geek/RenderEmbed}.
\end{abstract}

%% file: sections/1_intro.tex
\section{Introduction}

\begin{figure*}[t]
    \centering
    \includegraphics[width=\textwidth]{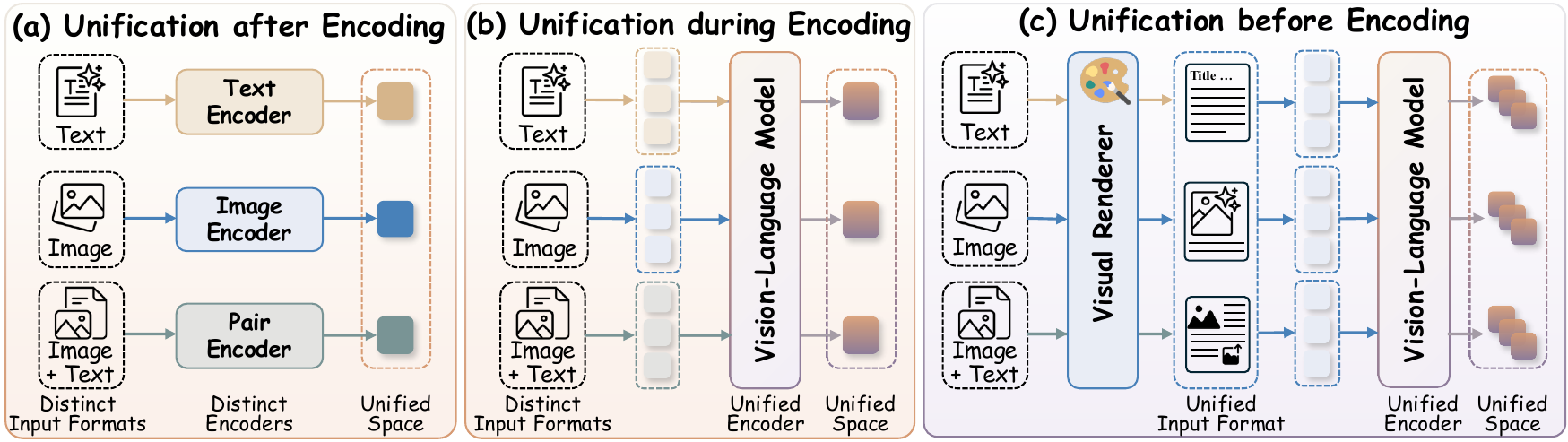}
    \caption{Paragigms of unified multimodal embedding.
    (a) \textbf{Unification after encoding} retains distinct input formats and encoders, aligning their outputs in a shared embedding space.
    (b) \textbf{Unification during encoding} introduces a shared VLM while retaining distinct input formats.
    (c) \textbf{Unification before encoding (Ours)}, further unifies the input format by converting all inputs into a common visual substrate. The stacked outputs in (c) denote compact multi-vector embeddings produced by our method.}
    \label{fig:motivation}
    \vspace{-1em}
\end{figure*}

Multimodal embedding models map heterogeneous inputs that may contain text, images, or combinations of both into a unified representation space, providing a common similarity interface for diverse tasks such as cross-modal search~\citep{radford2021learning,li2022blip,zhang2024badcm,he2026cross}, visual question answering~\citep{jiang2024vlm2vec,meng2026vlmvecv,jiang2026superglasses,jiang2026qadragon}, and retrieval-augmented generation~\citep{yu2025visrag,yuan2026mkg,yuan2026when,yuan2026ai,ding2026fashionkg}. 
Existing embedding methods achieve this unification through two main paradigms, as illustrated in Figure~\ref{fig:motivation}(a) and (b). Early approaches use modality-specific encoders and align their outputs in a shared embedding space~\citep{radford2021learning,jia2021scaling,li2022blip}. Recent universal embedding models use a shared vision-language model (VLM) to support diverse modality configurations~\citep{jiang2024vlm2vec,lin2025mm,zhang2024gme,li2026qwen3}. 
Within these models, text enters through a token embedding layer, while images pass through a vision encoder and projector before reaching the shared language backbone. The backbone must therefore accommodate modality-dependent input representations and their combinations. These distinct input pathways prevent heterogeneous tasks from jointly optimizing the same end-to-end encoding process, limiting the transfer of supervision across input configurations.

To unify this process, we explore \emph{unification before encoding} by expressing heterogeneous inputs in a common format before they enter the model. Recent work suggests that visual form can serve this role. Vision-based document retrievers encode pages directly as images, preserving textual, visual, and layout elements within a single visual input~\citep{ma2024unifying,faysse2025colpali}. In parallel, several studies render text into images for context compression~\citep{wei2025deepseek,cheng2026glyph} and visual latent reasoning~\citep{wang2026render,jiang2026univlr}, demonstrating that linguistic content can also be processed in visual form. Together, they suggest that different modalities can share a common visual substrate. This leads to our central question: \emph{can unified multimodal embedding be realized by adopting visual substrate as a common input format before encoding?}

To answer this question, we introduce \textbf{RenderEmbed}, a unified multimodal embedding framework built on \emph{unification before encoding}. As shown in Figure~\ref{fig:motivation}(c), RenderEmbed renders each input into a common visual substrate and encodes the resulting visual tokens into retrieval representations with a VLM. By standardizing the input format, it enables supervision from each modality configuration to optimize the same visual encoding pathway used by the others.

Implementing \emph{unification before encoding} requires a coherent path from visual construction to retrieval encoding and representation learning. Figure~\ref{fig:renderembed_framework} illustrates how RenderEmbed builds this path. RenderEmbed constructs a common visual substrate from heterogeneous inputs through \emph{Visual Substrate Rendering}, using a canonical configuration selected by retrieval-guided policy search. After rendering, the framework faces two challenges. First, it must encode retrieval evidence distributed across textual, visual, and layout structure into compact representations. It therefore introduces \emph{Progressive Hierarchical Anchor Encoding}, which uses progressively interleaved anchors to aggregate increasingly broad contexts and cross-layer fusion to combine multi-level representations from different depths. Second, it must learn robust retrieval representations that remain sensitive to content changes while remaining stable across different renderings. RenderEmbed promotes this robustness with \emph{Interventional Rendering Learning} through complementary semantic and rendering interventions.

Comprehensive experiments on MMEB~\citep{jiang2024vlm2vec} and ViDoRe v2~\citep{mace2025vidore} benchmarks, spanning retrieval tasks over text, images, and visual documents, demonstrate the effectiveness of RenderEmbed. It outperforms prior state-of-the-art models at the same scale, with particularly strong generalization to out-of-domain tasks. Further experiments show better transfer across input configurations and a growing advantage over the native-input baseline as training data scales.

Our main contributions are highlighted as follows:
\begin{itemize}
    \item We propose the \emph{unification before encoding} paradigm for unified multimodal embedding, enabling diverse modality configurations to jointly optimize the same end-to-end visual encoding pathway.
    \item We develop RenderEmbed, combining progressive hierarchical anchor encoding and interventional rendering learning to produce compact and robust retrieval representations from a common visual substrate.
    \item RenderEmbed achieves state-of-the-art performance and strong out-of-domain generalization, with improved transfer across input configurations and increasing gains as more training data is used.
\end{itemize}


%% file: sections/2_related_work.tex
\section{Related Work}

\textbf{Universal Multimodal Embedding.}
Multimodal embedding methods differ in where they unify heterogeneous inputs along the encoding pipeline. Classical models use modality-specific encoders and align their outputs in a shared embedding space after encoding~\citep{radford2021learning,jia2021scaling,li2022blip,zhou2024vista,wu2026lineage,huang2026grading}. 
More recent universal retrievers, such as VLM2Vec~\citep{jiang2024vlm2vec}, GME~\citep{zhang2024gme}, MM-EMBED~\citep{lin2025mm}, and MoCa~\citep{chen2026moca}, adapt vision-language models~\citep{liu2023visual,wang2024qwen2,abdin2024phi3,yuan2024instruction} to encode these input types within a shared backbone and training objective. Despite this architectural unification, their encoder inputs remain modality dependent, taking the form of text tokens, visual tokens, or interleaved multimodal sequences. RenderEmbed moves unification to the input stage by converting every retrieval unit into a common visual substrate before encoding.

\textbf{Visualized Information Retrieval.}
Visualized retrieval represents content through rendered text, generated images, or document screenshots. CLIPPO~\citep{tschannen2023clippo} and Pixel Linguist II~\citep{yuan2026design} explore pixel-based text encoding. CLIPPO applies it to image--text retrieval and adopts task-specific fine-tuning for VQA. VisRet~\citep{wu2026visret} conducts image retrieval by synthesizing images for textual querie. For document retrieval, DSE~\citep{ma2024unifying} and ColPali~\citep{faysse2025colpali} encode page images without content extraction. UniSE~\citep{liu2025any} extends screenshot-based retrieval to diverse tasks while retaining text tokens for textual inputs. In contrast, RenderEmbed converts all query and candidate inputs into a common visual form, enabling a single jointly trained embedding model to support diverse tasks.

\begin{figure*}
    \includegraphics[width=\textwidth]{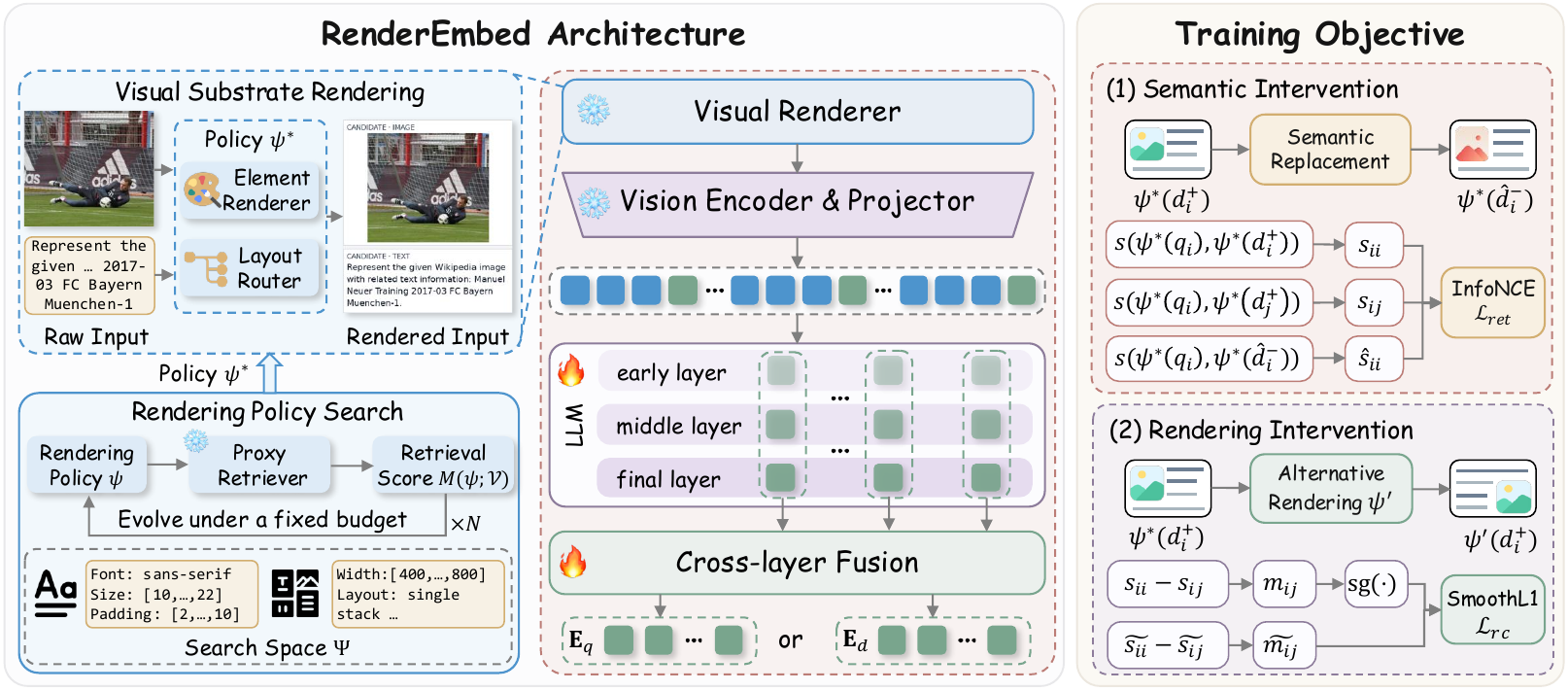}
    \caption{Overview of RenderEmbed. \textbf{Left}: Visual Substrate Rendering constructs a common visual substrate with a policy selected by retrieval-guided policy search. \textbf{Middle}: Progressive Hierarchical Anchor Encoding interleaves anchors with visual tokens and fuses their cross-layer states to produce compact retrieval vectors. \textbf{Right}: Interventional Rendering Learning combines semantic and rendering interventions to learn content-sensitive and rendering-robust retrieval representations.}
    \label{fig:renderembed_framework}
\end{figure*}

\textbf{Multi-Vector Retrieval.}
Multi-vector retrievers based on the late-interaction schema preserve fine-grained token- or patch-level evidence but increase index size and scoring cost~\citep{khattab2020colbert,santhanam2022colbertv2,faysse2025colpali,fan2025towards}. Retrieval-oriented compression reduces this overhead through token pooling, sequence compression, and patch pruning or merging~\citep{clavie2024reducing,ma2025towards,yan2026sculpting,yuan2026magic}. Recent approaches learn a small set of tokens to summarize inputs into compact multi-vector representations, as exemplified by AGC~\citep{qin2026multi} and MetaEmbed~\citep{xiao2026metaembed}. Within this paradigm, RenderEmbed uses Progressive Hierarchical Anchor Encoding to aggregate retrieval-relevant evidence from rendered inputs via progressive anchor interleaving and cross-layer fusion.

%% file: sections/3_method.tex
\section{Methodology}
\label{sec:method}

To realize \emph{unification before encoding} for unified multimodal embedding, we propose RenderEmbed, which transforms modality-dependent input formats into a shared visual substrate. This design enables training examples with diverse modality configurations to supervise a single end-to-end visual encoding pathway. As illustrated in Figure~\ref{fig:renderembed_framework}, RenderEmbed comprises three components: \textit{Visual Substrate Rendering} constructs unified visual inputs, \textit{Progressive Hierarchical Anchor Encoding} produces compact multi-vector representations, and \textit{Interventional Rendering Learning} learns robust representations for retrieval. We describe each component in the following sections.

\subsection{Problem Formulation}
\label{sec:method_formulation}

We consider a general multimodal retrieval setting, where a query $q$ is matched against a candidate collection $\mathcal{C}$. Both queries and candidates may contain text, images, or ordered compositions of both. Let $s(q,d)$ denote the relevance score between $q$ and a candidate $d \in \mathcal{C}$. Thus, the top-1 candidate selected by the retrieval model is determined by the highest relevance score: $d^\star = \arg\max_{d \in \mathcal{C}} s(q,d)$. 
For subsequent operations shared by both query and candidate sides, we use \(x\) to denote a generic retrieval unit representing either input.


\subsection{Visual Substrate Rendering}
\label{sec:method_rendering}

To establish a common input format before encoding, we introduce \emph{Visual Substrate Rendering}, which maps each retrieval unit $x$ to a common visual form $\psi(x)$ under a rendering policy $\psi$. The policy specifies the visual style, layout geometry, and element-routing rules. The rendering process consists of two stages. First, an \textit{Element Renderer} rasterizes text using the typography and style specified by $\psi$ and proportionally fits images into the designated visual regions. An \textit{Adaptive Layout Router} then assembles the rendered elements on a single pixel canvas while preserving their original order. In this way, these operations express retrieval units with different modality configurations in the same visual input format while preserving their content structure.

The common visual substrate is not unique: the same retrieval unit may admit multiple valid renderings that organize retrieval-relevant evidence differently and incur different visual-token costs. Since these choices affect both retrieval quality and encoding efficiency, we select a canonical policy $\psi^*$ from the candidate space $\Psi$ using validation data $\mathcal V$, subject to an average visual-token budget $\bar{\ell}$:
\begin{equation}
\label{eq:render_policy_selection}
\psi^*
=
\arg\max_{\psi\in\Psi}
\mathcal{M}(\psi;\mathcal V),
\ \mathrm{s.t.} \
\mathbb{E}_{x\sim\mathcal V}
\left[\ell\bigl(\psi(x)\bigr)\right]
\leq \bar{\ell}.
\end{equation}
Here, $\mathcal{M}$ measures retrieval performance, and $\ell(\cdot)$ counts visual tokens. We solve this discrete problem through evolutionary search~\citep{cheng2026glyph}, using MoCa-3B~\citep{chen2026moca} as a frozen proxy retriever, thereby decoupling rendering-policy selection from RenderEmbed optimization. Once selected, $\psi^*$ is fixed as the canonical policy throughout training and inference processes. Further details about the rendering process and policy search are provided in \textbf{Appendix~A}.

\subsection{Progressive Hierarchical Anchor Encoding}
\label{sec:method_encoding}

Rendered inputs distribute retrieval evidence across spatially separated text and image regions. Single-vector encoding can obscure local correspondences, whereas patch-level retrieval incurs high encoding, indexing, and scoring costs~\citep{faysse2025colpali}. We therefore seek a compact multi-vector representation~\citep{xiao2026metaembed} that maintains broad coverage under a fixed retrieval-vector budget. To this end, we propose \emph{Progressive Hierarchical Anchor Encoding}, which uses a small set of learnable anchors to aggregate progressively expanding contexts across the rendered visual sequence and applies cross-layer fusion to combine their multi-level representations.

RenderEmbed uses a pretrained Vision-Language Model (VLM) to encode each rendered retrieval unit. Its vision encoder and projector map the rendered image input $\psi(x)$ to a visual-token sequence:
\begin{equation}
\mathbf{E}_x
=
\mathrm{Projector}
\left(
\mathrm{VisEncoder}\bigl(\psi(x)\bigr)
\right)
\in \mathbb{R}^{N_x\times H},
\end{equation}
where $N_x$ is the number of visual tokens and $H$ is the hidden dimension of the VLM.
To obtain role-specific retrieval representations, we introduce separate learnable anchor banks for the query and candidate sides:
\begin{equation}
\begin{aligned}
\mathbf{A}_q
& =
[\mathbf{a}_{q,1},\ldots,\mathbf{a}_{q,M}] \in \mathbb{R}^{M\times H}, \\
\mathbf{A}_d
& =
[\mathbf{a}_{d,1},\ldots,\mathbf{a}_{d,M}] \in \mathbb{R}^{M\times H},
\end{aligned}
\end{equation}
where $M$ is the number of anchors. The two banks are learned independently, while all remaining encoding modules are shared. In what follows, $\mathbf{A}_x$ denotes the role-specific bank associated with $x$.

Appending all anchors after $\mathbf{E}_x$, as in MetaEmbed~\citep{xiao2026metaembed}, would give them the same complete visual prefix without explicitly differentiating their visual coverage. We instead partition $\mathbf{E}_x$ as evenly as possible into $M$ consecutive segments,
$\mathbf{E}_x=\mathbf{E}_{x,1}\Vert\cdots\Vert\mathbf{E}_{x,M}$,
and interleave one anchor after each segment:
\begin{equation}
\mathbf{Z}_x
=
\mathbf{E}_{x,1}\Vert\mathbf{a}_{x,1}
\Vert\cdots\Vert
\mathbf{E}_{x,M}\Vert\mathbf{a}_{x,M} \in \mathbb{R}^{(N_x+M)\times H},
\end{equation}
where $\Vert$ denotes sequence concatenation. The language backbone of VLM jointly processes the visual tokens and anchors in $\mathbf{Z}_x$. Under causal attention, the $i$-th anchor can attend to the first $i$ visual segments and all preceding anchors, so successive anchors aggregate nested and progressively expanding visual contexts.

Progressive interleaving expands each anchor's spatial scope, but its final-layer state reflects only one representational depth. Because fine-grained visual cues and contextual semantics are emphasized at different depths of the VLM, we fuse each anchor's hidden states across multiple layers. Concretely, let $\mathbf{h}_{x,i}^{(l)}$ denote the hidden state of anchor $\mathbf{a}_{x,i}$ at layer $l$. We collect its states from a set of $K$ selected layers, $\mathcal I_{\mathrm{fuse}} = \{l_1,\ldots,l_K\}$, which always includes the final layer. After applying RMSNorm~\citep{zhang2019root} to each state, we concatenate them and use a two-layer MLP with GELU activation to project the resulting $KH$-dimensional representation back to the $H$-dimensional retrieval space:
\begin{equation}
\label{eq:cross_layer_fusion}
\begin{aligned}
\widetilde{\mathbf{h}}_{x,i}^{(l)}
&= \operatorname{RMSNorm}
   \bigl(\mathbf{h}_{x,i}^{(l)}\bigr), \\
\mathbf{f}_{i}\bigl(\psi(x)\bigr)
&= \operatorname{MLP}\!\left(
   \left[
   \widetilde{\mathbf{h}}_{x,i}^{(l_1)};
   \ldots;
   \widetilde{\mathbf{h}}_{x,i}^{(l_K)}
   \right]\right).
\end{aligned}
\end{equation}
We then apply $\ell_2$ normalization to obtain the retrieval vector
$\mathbf{e}_{i}(\psi(x))
=
\mathbf{f}_{i}(\psi(x))/
\|\mathbf{f}_{i}(\psi(x))\|_2$.

Under the canonical rendering policy $\psi^*$, the VLM encodes $\psi^*(q)$ and $\psi^*(d)$ in separate forward passes, producing $M$ retrieval vectors for each. Their relevance score is computed using MaxSim operation in late interaction~\citep{khattab2020colbert}:
\begin{equation}
\label{eq:maxsim}
s\bigl(\psi^*(q),\psi^*(d)\bigr)
=
\sum_{i=1}^{M}
\max_{1\leq j\leq M}
\mathbf{e}_{i}\bigl(\psi^*(q)\bigr)^{\top}
\mathbf{e}_{j}\bigl(\psi^*(d)\bigr).
\end{equation}

\input{tables/mmeb_results}

\subsection{Interventional Rendering Learning}
\label{sec:method_training}

Rendering introduces presentation variations in addition to semantic content. A retrieval model should respond to changes in retrieval-relevant content while remaining robust when the same content is presented under different renderings. We therefore propose \emph{Interventional Rendering Learning}, which targets these complementary behaviors through a rendering-controlled semantic intervention and a semantics-preserving rendering intervention.

\paragraph{Rendering-Controlled Semantic Intervention.}
For a training batch $\mathcal{B}=\{(q_i,d_i^+,d_i^-)\}_{i=1}^{B}$, each query $q_i$ is paired with a labeled positive $d_i^+$ and negative $d_i^-$. After visual rendering, the negative $d_i^-$ may differ from $d_i^+$ in both retrieval-relevant content and visual organization, entangling semantic and rendering changes. To isolate the semantic difference, we construct a rendering-controlled negative $\hat d_i^-$ by replacing one retrieval-relevant element in $d_i^+$ with a same-type element from $d_i^-$ that contains different content. The intervention preserves the original layout geometry and unchanged regions. For image–text candidates whose relevance depends on both components, retaining one correct component while replacing the other encourages the model to distinguish fully relevant pairs from partial matches. \textbf{Appendix~B.2} details the construction and validation process.

Given the canonical policy $\psi^*$, let $s_{ij}=s\bigl(\psi^*(q_i),\psi^*(d_j^+)\bigr)$ denote the score between query $q_i$ and candidate $d_j^+$, and let $\hat s_{ii}=s\bigl(\psi^*(q_i),\psi^*(\hat d_i^-)\bigr)$ denote the score of the rendering-controlled negative. Standard InfoNCE contrasts the positive score $s_{ii}$ against $\{s_{ij}\}_{j\neq i}$, where the positives of other queries serve as in-batch negatives~\citep{jiang2024vlm2vec}. To encourage the model to distinguish retrieval-relevant content changes under matched rendering conditions, we add the controlled-negative score $\hat{s}_{ii}$ to the same denominator:
\begin{equation}
\label{eq:retrieval_loss}
\mathcal{L}_{\mathrm{ret}}
=
-\frac{1}{B}
\sum_{i=1}^{B}
\log
\frac{
\exp(s_{ii}/\tau)
}{
\sum_{j=1}^{B}
\exp(s_{ij}/\tau)
+
\exp(\hat s_{ii}/\tau)
},
\end{equation}
where $\tau$ denotes the temperature.

\paragraph{Semantics-Preserving Rendering Intervention.}
The semantic intervention encourages sensitivity to retrieval-relevant content changes, while the complementary objective is to preserve retrieval decisions across alternative renderings of the same content. We sample a policy $\psi' \neq \psi^*$ and use it to render each positive candidate $d_j^+$, varying its presentation while preserving its content and element order. Pairing the alternative candidate rendering $\psi'(d_j^+)$ with the canonical query rendering $\psi^*(q_i)$ yields the alternative-rendering score $\widetilde{s}_{ij}=s\bigl(\psi^*(q_i),\psi'(d_j^+)\bigr)$.

Alternative renderings can produce different visual-token sequences and retrieval vectors for the same content, making direct embedding alignment overly restrictive. Thus, we align the relative retrieval decisions induced by the canonical and alternative renderings. For $i \neq j$, their positive-to-negative margins are defined as
$m_{ij}=s_{ii}-s_{ij}$
and
$\widetilde{m}_{ij}=\widetilde{s}_{ii}-\widetilde{s}_{ij}$.
Their rendering consistency is then optimized by
\begin{equation}
\label{eq:rendering_consistency}
\mathcal{L}_{\mathrm{rc}}
=
\frac{1}{B(B-1)}
\sum_{i=1}^{B}
\sum_{j\neq i}
\operatorname{SmoothL1}
\left(
\widetilde{m}_{ij}
-
\operatorname{sg}
\left(
m_{ij}
\right)
\right),
\end{equation}
where $\operatorname{sg}(\cdot)$ denotes stop-gradient. The canonical margin serves as the reference, allowing the retrieval vectors to vary across rendering settings while preserving the relative retrieval structure. At each training step, $\psi'$ is resampled from a pool of alternative policies, as described in \textbf{Appendix~B.2}.

\paragraph{Training Objective.}
The final objective combines the intervention-augmented contrastive loss with rendering-consistency regularization: $\mathcal{L} = \mathcal{L}_{\mathrm{ret}} + \lambda_{\mathrm{rc}}\mathcal{L}_{\mathrm{rc}}$, where $\lambda_{\mathrm{rc}}$ balances the two objectives. Both interventions are confined to training, while inference use the canonical policy $\psi^*$ without controlled negatives or alternative rendering views.


%% file: tables/mmeb_results.tex
\begin{table*}[t]
\centering
\caption{Main results on MMEB. Precision@1 is reported for each model. IND and OOD denote in-domain and out-of-domain averages.}
\label{tab:mmeb_results}
\resizebox{0.92\linewidth}{!}{
\begin{tabular}{lccccccccc}
\toprule
\multirow{2}{*}{\textbf{Model}} & \multirow{2}{*}{\textbf{Venue}} & \multirow{2}{*}{\textbf{Size}} &
\multicolumn{4}{c}{\textbf{Per Meta-Task Score}} & \multicolumn{3}{c}{\textbf{Average Score}} \\
\cmidrule(lr){4-7} \cmidrule(lr){8-10}
 &  &  & Classification & VQA & Retrieval & Grounding & IND & OOD & Overall \\
\midrule
CLIP~\citep{radford2021learning}       & ICML'21 & 428M & 42.8 & 9.1  & 53.0 & 51.8 & 37.1 & 38.7 & 37.8 \\
BLIP2~\citep{li2023blip}               & ICML'23 & 428M & 27.0 & 4.2  & 33.9 & 47.0 & 25.3 & 25.1 & 25.2 \\
OpenCLIP~\citep{cherti2023reproducible}& CVPR'23 & 428M & 47.8 & 10.9 & 52.3 & 53.3 & 39.3 & 40.2 & 39.7 \\
SigLIP~\citep{zhai2023sigmoid}         & ICCV'23 & 652M & 40.3 & 8.4  & 31.6 & 59.5 & 32.3 & 38.0 & 34.8 \\
Magiclens~\citep{zhang2024magiclens}   & ICML'24 & 613M & 38.8 & 8.3  & 35.4 & 26.0 & 31.0 & 23.7 & 27.8 \\
UniIR~\citep{wei2024uniir}             & ECCV'24 & 428M & 42.1 & 15.0 & 60.1 & 62.2 & 44.7 & 40.4 & 42.8 \\
GME~\citep{zhang2024gme}               & CVPR'25 & 7B   & 58.6 & 35.8 & 73.1 & 67.3 & 55.8 & 60.9 & 58.0 \\
VLM2Vec~\citep{jiang2024vlm2vec}       & ICLR'25 & 7B   & 61.2 & 49.9 & 67.4 & 86.1 & 67.5 & 57.1 & 62.9 \\
VLM2Vec-V2~\citep{meng2026vlmvecv}     & TMLR'26 & 2B   & 62.9 & 56.3 & 69.5 & 77.3 & --   & --   & 64.9 \\
MM-Embed~\citep{lin2025mm}             & ICLR'25 & 7B   & 48.1 & 32.2  & 63.8 & 57.8 & --  & --   & 50.0 \\
MMRet~\citep{zhou2025megapairs}        & ACL'25  & 7B   & 56.0 & 57.4 & 69.9 & 83.6 & 68.0 & 59.1 & 64.1 \\
UniME-V2-2B~\citep{gu2026unime}        & AAAI'26 & 2B & 49.5 & 52.0 & 58.0 & 65.9 & 59.2 & 49.5 & 54.9 \\
UniME-V2-7B~\citep{gu2026unime}        & AAAI'26 & 7B & 51.9 & 56.4 & 62.8 & 74.3 & 62.3 & 55.5 & 59.3 \\
MoCa-3B~\citep{chen2026moca}           & ACL'26  & 3B   & 59.8 & 62.9 & 70.6 & \textbf{88.6} & 72.3 & 61.5 & 67.5 \\
MetaEmbed~\citep{xiao2026metaembed}    & ICLR'26 & 3B   & 62.7 & 68.1 & 71.9 & 79.6 & 73.5 & 63.8 & 69.1 \\
\midrule
\rowcolor{gray!15} RenderEmbed (Ours)  & --      & 3B   & \textbf{64.3} & \textbf{68.9} & \textbf{73.6} & 83.2 & \textbf{74.7} & \textbf{65.9} & \textbf{70.8} \\
\bottomrule
\end{tabular}
}
\vspace{-0.5em}
\end{table*}

%% file: sections/4_experiment.tex
\section{Experiments}
\label{sec:experiments}

\input{tables/vidore_results}

\subsection{Experimental Setup}
\label{sec:exp_setup}

\paragraph{Training Data and Benchmarks.} Following the experimental setting of MetaEmbed~\citep{xiao2026metaembed}, we train RenderEmbed on MoCa-CL-Pairs~\citep{chen2026moca}, which combines the training set of MMEB~\citep{jiang2024vlm2vec} and ViDoRe~\citep{faysse2025colpali}. MMEB contains 20 datasets spanning four meta-tasks: classification, visual question answering (VQA), multimodal retrieval, and visual grounding~\citep{jiang2024vlm2vec}. ViDoRe focuses on visually rich document retrieval, covering documents from public academic datasets and web-crawled PDFs. We evaluate general multimodal embedding on MMEB using Precision@1 and visual document retrieval on ViDoRe v2~\citep{faysse2025colpali} using average NDCG@5. The MMEB comparison includes multimodal embedding models, whereas ViDoRe v2 additionally includes multi-vector models designed for visual document retrieval.

\paragraph{Implementation Details.} RenderEmbed is initialized from Qwen2.5-VL-3B~\citep{bai2025qwen25vl}. We train the role-specific anchor banks and the cross-layer fusion module. Every LLM layer is adapted with LoRA~\citep{hu2022lora} using rank 32 and scaling factor \(\alpha=32\), while all remaining parameters are frozen. The final configuration uses \(M=\mathrm{16}\) anchors and \(\mathcal I_{\mathrm{fuse}}=\{\mathrm{18, 24, 30, 36}\}\). Training runs for \(\mathrm{1}\) epoch with a global batch size of \(\mathrm{128}\), temperature \(\tau=\mathrm{0.02}\), and rendering-consistency weight \(\lambda_{\mathrm{rc}}=\mathrm{0.1}\), while each query-positive pair contributes one rendering-controlled semantic negative and one alternative rendering view. For RenderEmbed, all training and evaluation inputs pass through Visual Substrate Rendering before encoding. More experimental details are provided in \textbf{Appendix~B}.

\subsection{Main Results}
\label{sec:exp_main_results}

Tables~\ref{tab:mmeb_results} and~\ref{tab:vidore_results} summarize the main results on general multimodal embedding and visual document retrieval, respectively. RenderEmbed, MoCa-3B, ColQwen2.5, and MetaEmbed share the same Qwen2.5-VL-3B backbone, enabling capacity-matched comparisons across retrieval representation types. Specifically, MoCa-3B uses a single-vector representation, ColQwen2.5 retains patch-level vectors, and MetaEmbed learns a compact multi-vector representation. Our RenderEmbed instead constructs compact multi-vector embeddings from a common visual substrate, using \(M=\mathrm{16}\) vectors on both the query and candidate sides.

\paragraph{General Multimodal Embedding.} 
Across MMEB, RenderEmbed achieves the strongest overall performance, leading three of the four meta-tasks as well as both IND and OOD averages. It outperforms the state-of-the-art 3B baselines MetaEmbed and MoCa-3B under the same backbone, alongside all compared 7B models in Table~\ref{tab:mmeb_results}. Its advantage over both 3B baselines is larger on OOD tasks, reaching 2.1 points over MetaEmbed and 4.4 points over MoCa-3B. These larger OOD gains suggest that the shared visual pathway facilitates transfer beyond the training tasks. Rendering heterogeneous inputs into a common format allows supervision from different tasks to refine the same input-to-retrieval mapping, encouraging the reuse of retrieval knowledge across input configurations. Section~\ref{sec:exp_analysis} examines this transfer in more detail, while \textbf{Appendix~C.3} evaluates visual rendering in the single-vector VLM2Vec architecture.

\paragraph{Visual Document Retrieval.} 
The performance advantage also extends to visual document retrieval: on ViDoRe v2, RenderEmbed obtains the highest average NDCG@5 among the compared single-vector and multi-vector methods and outperforms all three capacity-matched baselines. Its strongest results appear on multilingual retrieval, where it ranks first on the multilingual Biomedical and ESG subsets despite the limited share of multilingual examples in the training corpus. By rendering multilingual inputs into the same visual substrate as the rest of the training corpus, RenderEmbed allows limited multilingual supervision to update a common visual encoding pathway, which may facilitate transfer across languages. \textbf{Appendix~C.4} further evaluates visual-document and video tasks in MMEB-V2.

\subsection{Ablation Studies}
\label{sec:exp_ablation}

\input{tables/ablation}

\paragraph{Core Design.} Table~\ref{tab:ablation} compares variants using the same 25\% training subset, backbone, and optimization settings. To assess unified visual inputs, we compare RenderEmbed with NativeEmbed, which uses native inputs and standard hard negatives. RenderEmbed improves Overall Precision@1 by 1.1 points. Interventional Rendering Learning further exploits visual inputs by distinguishing content changes from rendering variations. Both interventions reuse existing training triples without additional source data or annotations. Replacing these objectives with contrastive training using standard hard negatives (\textit{w/o Interventional Rendering Learning}) lowers accuracy by 0.5 points.

Within the encoder architecture, appending all anchors after the input and using only final-layer states reduce accuracy by 0.4 and 0.7 points, respectively. These results suggest that compact retrieval vectors benefit from diverse context coverage and information from multiple backbone depths. For the learning objectives, \textit{w/o Semantic Intervention} retains rendering consistency but replaces controlled negatives with standard hard negatives, reducing accuracy by 0.5 points. Removing rendering consistency has a smaller effect of 0.2 points, leaving accuracy at 68.4. This remains above the 67.9 achieved by ordinary rendering augmentation, whose setup is described in \textbf{Appendix~C.1}. Rendering consistency primarily targets robustness to presentation changes, which we examine in Section~\ref{sec:exp_analysis}.

\begin{figure}[t]
    \centering
    \begin{minipage}[t]{0.54\linewidth}
        \centering
        \includegraphics[width=\linewidth]{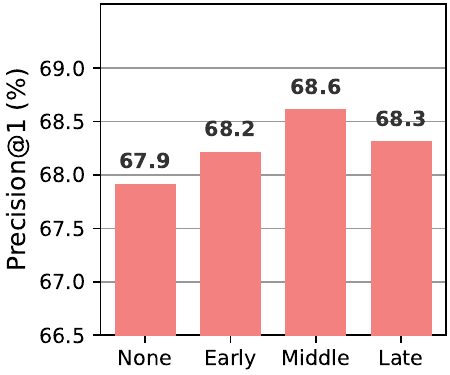}
        \small (a) Fusion Layers
    \end{minipage}\hfill
    \begin{minipage}[t]{0.42\linewidth}
        \centering
        \includegraphics[width=\linewidth]{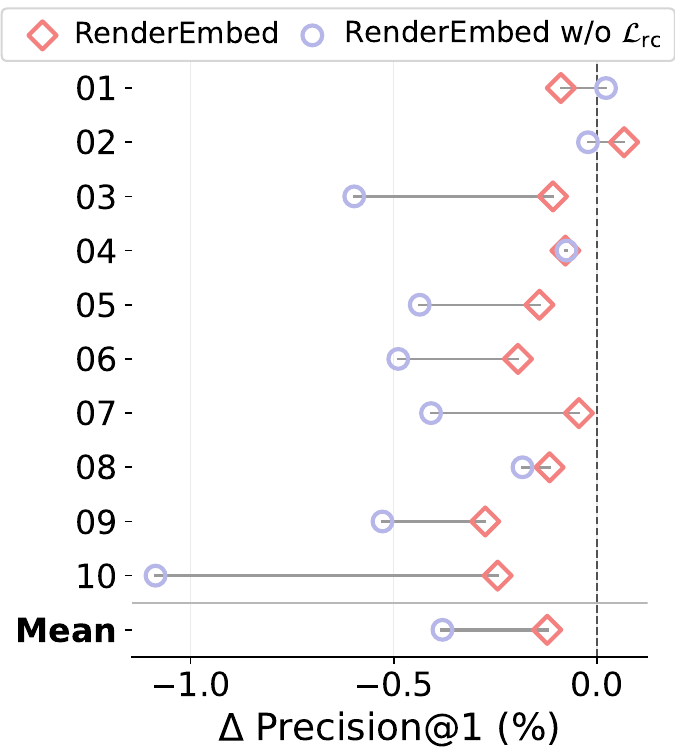}
        \small (b) Rendering Sensitivity
    \end{minipage}
    \caption{Fusion-layer ablation and rendering sensitivity on MMEB. (a) Overall Precision@1 under different fusion-layer settings. (b) Change in Overall Precision@1 for RenderEmbed and \textit{w/o Rendering Intervention} variant under 10 different non-canonical rendering policies. Changes are measured relative to the canonical rendering policy, with positive and negative values indicating improvements and degradations, respectively. The vertical dashed line denotes no change, and the \textit{Mean} row gives the average across policies.}
    \label{fig:ablation_fusion}
\end{figure}

\paragraph{Fusion Layers.}
We further analyze which backbone depths contribute most effectively to cross-layer fusion. For the 36-layer Qwen2.5-VL-3B backbone, we compare early fusion with \(\mathcal I_{\mathrm{fuse}}=\{6,12,18,36\}\), middle fusion with \(\{18,24,30,36\}\), and late fusion with \(\{33,34,35,36\}\). All settings fuse four layer states and include the final layer, while the no-fusion baseline uses only \(\{36\}\). Figure~\ref{fig:ablation_fusion}(a) shows that all cross-layer configurations outperform using the final layer alone, with middle fusion performing best. This pattern suggests that effective fusion requires representations that are sufficiently deep to capture retrieval-relevant semantics while remaining separated enough to provide complementary information.

\begin{figure}[t]
    \centering
    \begin{minipage}[t]{0.49\linewidth}
        \centering
        \includegraphics[width=\linewidth]{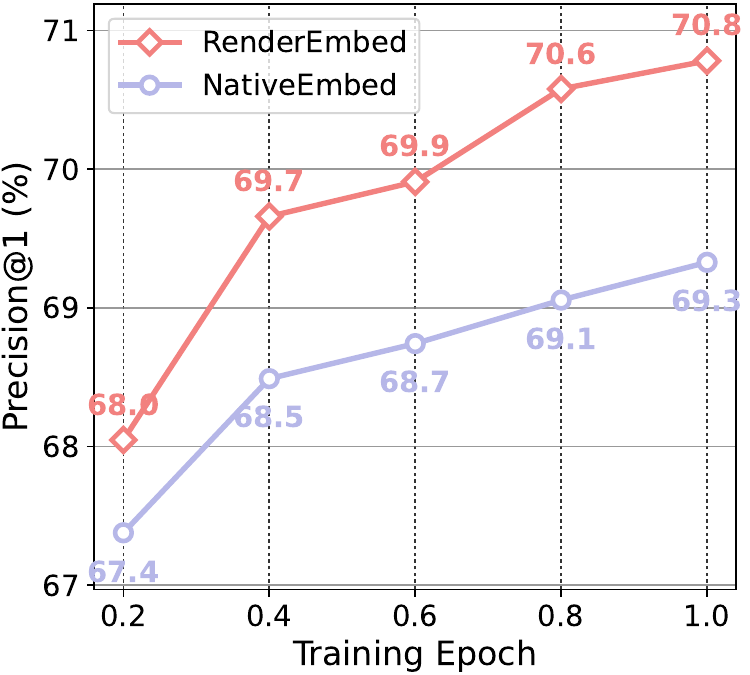}
        \small (a) Training-Data Scaling
    \end{minipage}\hfill
    \begin{minipage}[t]{0.49\linewidth}
        \centering
        \includegraphics[width=\linewidth]{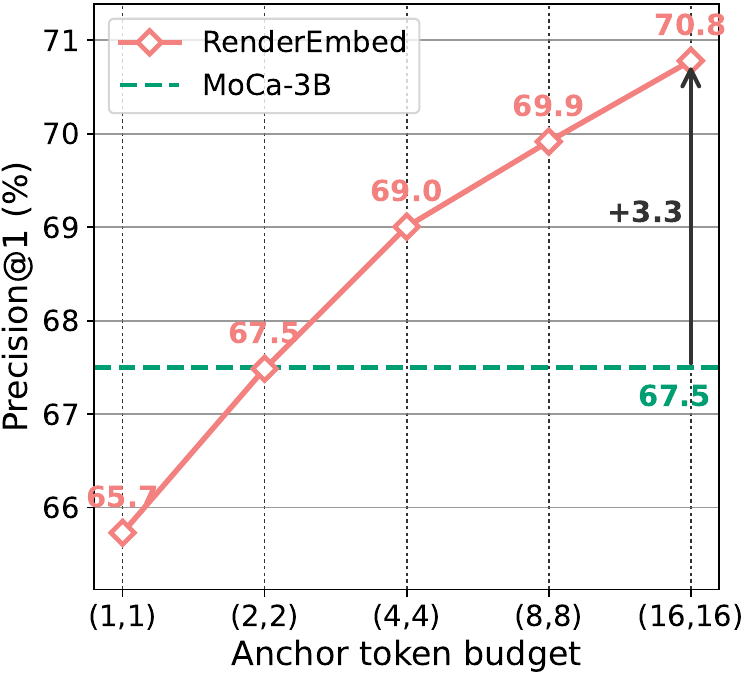}
        \small (b) Test-Time Anchor Budget
    \end{minipage}
    \caption{(a) Overall MMEB Precision@1 of RenderEmbed and NativeEmbed with various training epochs, representing increasing training budgets. (b) Overall MMEB Precision@1 under different anchor budgets. The dashed line marks the MoCa-3B baseline.}
    \label{fig:analysis}
\end{figure}

\subsection{Analysis}
\label{sec:exp_analysis}

\paragraph{Training-Data Scaling.}
Figure~\ref{fig:analysis}(a) evaluates MMEB Overall Precision@1 after training for 0.2, 0.4, 0.6, 0.8, and 1.0 epochs. Both methods improve with larger training budgets, while RenderEmbed consistently outperforms NativeEmbed at every budget. The advantage also increases from 0.6 points at 0.2 epochs to 1.5 points at 1.0 epoch, indicating that RenderEmbed benefits increasingly from additional training data. We attribute this trend to the the proposed \emph{unification before encoding} paradigm, which allows heterogeneous training corpus to jointly optimize the same end-to-end visual encoding pathway and thereby improves the utilization of additional training data.

\paragraph{Transfer Across Input Configurations.}
To assess whether a shared visual pathway facilitates supervision transfer, we train NativeEmbed and RenderEmbed on $T$ or $T+S$. The target group $T$ contains 25K examples each from MSCOCO and VisualNews for text-to-image retrieval. The auxiliary group $S$ adds 25K CIRR examples for composed text--image-to-image retrieval. NIGHTS evaluates image-to-image retrieval and is held out in both settings. 

With $T$ alone, RenderEmbed shows mixed results on the target tasks but gains 0.9 points over NativeEmbed on both CIRR and NIGHTS, as illustrated in Table~\ref{tab:transfer}, supporting transfer to unseen input configurations. This transfer advantage grows with heterogeneous supervision: adding $S$ improves held-out NIGHTS accuracy by 2.3 points for RenderEmbed and 1.7 points for NativeEmbed. On VisualNews, the same addition lowers NativeEmbed's accuracy by 0.6 points but raises RenderEmbed's by 0.5 points, suggesting that the shared visual pathway also helps mitigate interference between input configurations.

\input{tables/transfer}

\paragraph{Test-Time Anchor Budget.}
Figure~\ref{fig:analysis}(b) evaluates RenderEmbed under symmetric test-time anchor budgets, retaining the same number of query and candidate anchors. MetaEmbed learns prefix-truncatable representations through the Matryoshka Multi-Vector Retrieval objective~\citep{xiao2026metaembed}, whereas RenderEmbed supports test-time budget adaptation without an additional budget-specific training objective. For a target budget, we uniformly sample anchors across their progressively interleaved positions and always retain the final anchor. Because successive anchors aggregate progressively broader visual contexts, this strategy preserves multiple context scales while maintaining coverage of the complete input. 
Performance increases monotonically with the anchor budget: using two anchors per side already matches MoCa-3B while retaining only \(1/8\) of the full-budget candidate vectors and requiring \(1/64\) of the anchor-pair comparisons. With the full \((16,16)\) budget, RenderEmbed exceeds MoCa-3B by 3.3 points. This monotonic curve provides a flexible quality--efficiency trade-off from a single trained model. \textbf{Appendix~C.2} further shows that, once the query retains several anchors, candidate-heavy allocations achieve higher accuracy at matched scoring cost.

\paragraph{Rendering Sensitivity.}
To test robustness to presentation changes, we compare RenderEmbed with the \textit{w/o Rendering Intervention} under 10 non-canonical policies sampled from the policy candidate space, as shown in Figure~\ref{fig:ablation_fusion}(b). Under each policy, both queries and candidates are rendered accordingly, and we report the change in Overall Precision@1 relative to canonical rendering. Both variants remain relatively robust across these policies, with all performance changes bounded within 1.2 points, indicating that the underlying RenderEmbed framework is already tolerant to rendering variation. Across policies, the changes are \(-0.12\pm0.09\) points for RenderEmbed and \(-0.38\pm0.32\) for \textit{w/o Rendering Intervention} (mean \(\pm\) standard deviation). RenderEmbed shows less average degradation and lower variability, outperforming under 8/10 policies and exhibiting a smaller worst-case drop. By aligning positive-to-negative margins between canonical and alternative renderings, the rendering-consistency objective further stabilizes retrieval decisions across changes in layout and visual style.

\input{tables/efficiency}

\paragraph{Efficiency.}
In Table~\ref{tab:efficiency}, we compare RenderEmbed with representative multi-vector retrievers in terms of representation storage and late-interaction scoring efficiency. RenderEmbed performs best across all three efficiency metrics. MetaEmbed-3B provides the most direct comparison because it uses the same embedding dimension and query-vector budget as RenderEmbed. Compared with its 64 candidate vectors, the 16-vector candidate representation of RenderEmbed reduces index storage and scoring FLOPs by a factor of four, while lowering scoring latency from \(4.18\) to \(1.30\) ms, corresponding to a \(3.2\times\) speedup. \textbf{Appendix~B.3} provides the measurement protocol and further analyzes encoding and training overheads.

%% file: tables/vidore_results.tex
\begin{table*}[t]
\centering
\caption{Main results on ViDoRe v2. The metric is nDCG@5 (\%). ``EN'' is the English subset, ``Multi. Avg.'' denotes the average over multilingual queries, and ``Human'' denotes the human-labeled subset.}
\label{tab:vidore_results}
\resizebox{0.95\linewidth}{!}{
\begin{tabular}{lcc|cc|cc|ccc|c}
\toprule
\multirow{2}{*}{\textbf{Method}} &
\multirow{2}{*}{\textbf{Venue}} &
\multirow{2}{*}{\textbf{Size}} &
\multicolumn{2}{c|}{\textbf{Biomedical Lectures}} &
\multicolumn{2}{c|}{\textbf{Economics Reports}} &
\multicolumn{3}{c|}{\textbf{ESG Reports}} &
\multirow{2}{*}{\textbf{Avg.}} \\
\cmidrule(lr){4-5}\cmidrule(lr){6-7}\cmidrule(lr){8-10}
& & & EN & Multi. Avg. & EN & Multi. Avg. & EN & Multi. Avg. & Human & \\
\midrule
\rowcolor{gray!15} \multicolumn{11}{c}{~\textit{\textbf{Single-Vector Retrieval}}} \\
SigLIP~\citep{zhai2023sigmoid}        & ICCV'23 & 652M & 33.8 & 18.2 & 29.8 & 14.0 & 19.8 & 21.9 & 28.8 & 23.8 \\
VLM2Vec~\citep{jiang2024vlm2vec}      & ICLR'25 & 7B   & 38.8 & 29.7 & 51.4 & 42.0 & 36.7 & 38.4 & 33.9 & 38.7 \\
VisRAG-Ret~\citep{yu2025visrag}       & ICLR'25 & 3B   & 54.8 & 47.7 & 59.6 & 48.7 & 45.9 & 46.4 & 53.7 & 51.0 \\
GME~\citep{zhang2024gme}              & CVPR'25 & 7B   & 64.0 & 55.1 & \textbf{62.9} & 56.2 & 54.3 & 56.7 & \textbf{65.8} & 59.3 \\
MoCa-3B~\citep{chen2026moca}          & ACL'26  & 3B   & 62.5 & 59.8 & 62.8 & \textbf{57.3} & 58.3 & 54.8 & 63.3 & 59.8 \\
\midrule
\rowcolor{gray!15} \multicolumn{11}{c}{~\textit{\textbf{Multi-Vector Retrieval}}} \\
ColPali~\citep{faysse2025colpali}     & ICLR'25 & 3B   & 59.7 & 56.5 & 51.6 & 49.9 & 57.0 & 55.7 & 51.1 & 54.5 \\
ColQwen2.5~\citep{faysse2025colpali}  & ICLR'25 & 3B   & 63.2 & 60.3 & 61.0 & 57.2 & 60.6 & 56.2 & 64.6 & 60.4 \\
Jina-v4~\citep{gunther2025jina}       & MRL'25  & 4B   & 65.1 & 62.0 & 57.6 & 54.5 & 50.9 & 52.7 & 60.9 & 57.7 \\
AGC~\citep{qin2026multi}              & SIGIR'26 & 3B   & 60.2 & 57.7 & 62.8 & 56.3 & 54.5 & 53.1 & 54.5 & 57.0 \\
MAGIC~\citep{yuan2026magic}           & arXiv'26 & 3B   & 60.6 & 57.8 & 60.3 & 56.8 & 57.2 & 53.2 & 55.4 & 57.3 \\
MetaEmbed~\citep{xiao2026metaembed}   & ICLR'26 & 3B   & 61.7 & 58.7 & 62.3 & 55.5 & \textbf{62.6} & 57.4 & 63.7 & 60.3 \\
\midrule
\rowcolor{gray!15} RenderEmbed (Ours) & --      & 3B   & \textbf{66.2} & \textbf{62.1} & 59.9 & 55.6 & 60.6 & \textbf{58.2} & 63.6 & \textbf{60.9} \\
\bottomrule
\end{tabular}
}
\end{table*}

%% file: tables/ablation.tex
\begin{table}[t]
\centering
\caption{Ablation study on the core designs of RenderEmbed. $\Delta$ is the change in MMEB Overall Precision@1 relative to our RenderEmbed.}
\label{tab:ablation}
\resizebox{0.95\linewidth}{!}{%
\begin{tabular}{lcc}
    \toprule
    \textbf{Model} & \textbf{Overall $\uparrow$} & \textbf{$\Delta$} \\
    \midrule
    \rowcolor{gray!15} RenderEmbed                 & 68.6 & -- \\
    NativeEmbed                                    & 67.5 & \(-1.1\) \\
    \midrule
    \textit{w Appended Anchors}                    & 68.2 & \(-0.4\) \\
    \textit{w Final Layer Only}                    & 67.9 & \(-0.7\) \\
    \textit{w/o Semantic Intervention}             & 68.1 & \(-0.5\) \\
    \textit{w/o Rendering Intervention}            & 68.4 & \(-0.2\) \\
    \textit{w Rendering Augmentation}              & 67.9 & \(-0.7\) \\
    \textit{w/o Interventional Rendering Learning} & 68.1 & \(-0.5\) \\
    \bottomrule
\end{tabular}
}
\end{table}

%% file: tables/transfer.tex
\begin{table}[t]
\centering
\caption{Transfer across input configurations, measured by Precision@1 (\%). Avg. denotes the mean over the four datasets.}
\label{tab:transfer}
\setlength{\tabcolsep}{2pt}
\resizebox{\linewidth}{!}{
\begin{tabular}{ccccccc}
\toprule
\makecell[c]{\textbf{Training}\\\textbf{Data}} & \textbf{Model} & \makecell[c]{\textbf{MSCOCO}\\T2I} & \makecell[c]{\textbf{VisualNews}\\T2I} & \makecell[c]{\textbf{CIRR}\\TI2I} & \makecell[c]{\textbf{NIGHTS}\\I2I} & \textbf{Avg.} \\
\midrule
\multirow{2}{*}{$T$} & NativeEmbed & 75.8 & \textbf{73.9} & 18.2 & 64.4 & 58.1 \\
 & RenderEmbed & \textbf{76.1} & 73.2 & \textbf{19.1} & \textbf{65.3} & \textbf{58.4} \\
\midrule
\multirow{2}{*}{$T+S$} & NativeEmbed & 76.1 & 73.3 & 61.7 & 66.1 & 69.3 \\
 & RenderEmbed & \textbf{76.3} & \textbf{73.7} & \textbf{63.1} & \textbf{67.6} & \textbf{70.2} \\
\bottomrule
\end{tabular}
}
\end{table}

%% file: tables/efficiency.tex
\begin{table}[t]
\centering
\caption{Retrieval efficiency for 1,000 queries and 100,000 candidates. Q/C and Dim. denote average query/candidate vector counts and embedding dimension, respectively. Index size is the BF16 storage for all candidate embeddings. FLOPs and Latency measure per-query scoring operations and time over all candidates, excluding encoding. Latency is reported as \(\mathrm{mean}_{\pm\mathrm{std}}\) over 10 runs.}
\label{tab:efficiency}
\footnotesize
\setlength{\tabcolsep}{2.5pt}
\resizebox{\columnwidth}{!}{%
\begin{tabular}{lccccc}
\toprule
\makecell[c]{\textbf{Model}} &
\makecell[c]{\textbf{Q/C}\\\textbf{Vectors}} &
\makecell[c]{\textbf{Dim.}} &
\makecell[c]{\textbf{Index Size}\\\textbf{(GiB)}} &
\makecell[c]{\textbf{FLOPs}\\\textbf{(G)}} &
\makecell[c]{\textbf{Latency}\\\textbf{(ms)}} \\
\midrule
ColPali                         & 31 / 1031 & 128  & 24.58  & 826.17   & \(13.06_{\pm 7.0\times10^{-4}}\) \\
ColQwen2.5                      & 35 / 751  & 128  & 17.90  & 666.05   & \(11.58_{\pm 8.3\times10^{-4}}\) \\
MetaEmbed-3B                    & 16 / 64   & 2048 & 24.41  & 419.43   & \(4.18_{\pm 1.5\times10^{-4}}\) \\
\rowcolor{gray!15} RenderEmbed  & 16 / 16   & 2048 & \textbf{6.10}   & \textbf{104.85}   & \(\mathbf{1.30_{\pm 1.0\times10^{-4}}}\) \\
\bottomrule
\end{tabular}
}
\end{table}

%% file: sections/5_conclusion.tex
\section{Conclusion}
RenderEmbed demonstrates that multimodal unification can be extended to the input level through \emph{unification before encoding}. It uses \emph{Visual Substrate Rendering} to convert text, images, and their combinations into a common visual substrate, allowing heterogeneous supervision to optimize the same visual encoding pathway. The framework further integrates \emph{Progressive Hierarchical Anchor Encoding} to retain distributed evidence in compact retrieval representations and \emph{Interventional Rendering Learning} to distinguish retrieval semantics from presentation variations. Across evaluation benchmarks, RenderEmbed attains state-of-the-art performance at the same scale and strong out-of-domain generalization. Transfer and scaling gains further support the value of unified visual inputs.

\subsection*{LLM Usage Statement}

In this work, we used LLM tools to support the retrieval and discovery of related literature, assist with the implementation of auxiliary training and testing scripts, and aid in drafting and polishing the manuscript. LLM was not used to design or implement the core method and experiments. The authors reviewed all LLM-assisted work and take responsibility for the final content of this work, including its text, claims, code, and other artifacts produced with the aid of LLM.

%% file: sections/appendix.tex

We have included supplementary material to facilitate a more comprehensive understanding and in-depth analysis of the primary paper. The supplementary material is organized as follows:

\begin{itemize}
    \item \textbf{Section~\ref{app:rendering_details}: Rendering Implementation Details}
    \item \textbf{Section~\ref{app:experimental_details}: Additional Experimental Details}
    \item \textbf{Section~\ref{app:additional_results}: Additional Experimental Results}
    \item \textbf{Section~\ref{app:further_analysis}: Further Analysis}
\end{itemize}

\section{Rendering Implementation Details}
\label{app:rendering_details}

This section provides the implementation and selection details of Visual Substrate Rendering. Section~\ref{app:rendering_construction} describes how heterogeneous inputs, including text, image, and composed elements, are rendered into visual form and routed among the supported layout templates. Section~\ref{app:rendering_policy_search} defines the rendering-policy space  and presents the retrieval-guided policy search used to select a canonical rendering policy. Finally, Section~\ref{app:selected_rendering_policy} analyzes the search outcomes and specifies the selected canonical rendering policy.

\subsection{Element Rendering and Adaptive Layout Routing}
\label{app:rendering_construction}

Visual Substrate Rendering maps each retrieval unit \(x\) to a visual substrate \(\psi(x)\) under a rendering policy \(\psi\). A rendering policy is a parameterized specification of the complete rendering process: it determines how the text and image elements of \(x\) are visually instantiated and how the rendered elements are spatially composed into a common substrate. This process is implemented by two components. The Element Renderer converts individual elements into visual cards, while the Adaptive Layout Router selects among the supported layouts according to the input composition and rendered geometry.

\paragraph{Element Renderer.} For a retrieval unit \(x\), the Element Renderer retains its non-empty text and image elements in their original order and renders each element as a typed visual card. Each card contains a header that identifies its retrieval role (query or candidate) and element type (text or image), together with a content region, background, and border. The parameters in \(\psi\) determine the typography, color theme, and internal padding. Text is wrapped within the available content width, whereas an image is fitted proportionally to its assigned region. Given the assigned card width, the header, rendered content, and internal padding determine the final card height.

\input{tables/appendix/layout_template.tex}

\begin{figure*}[t]
    \centering
    \includegraphics[width=0.9\linewidth]{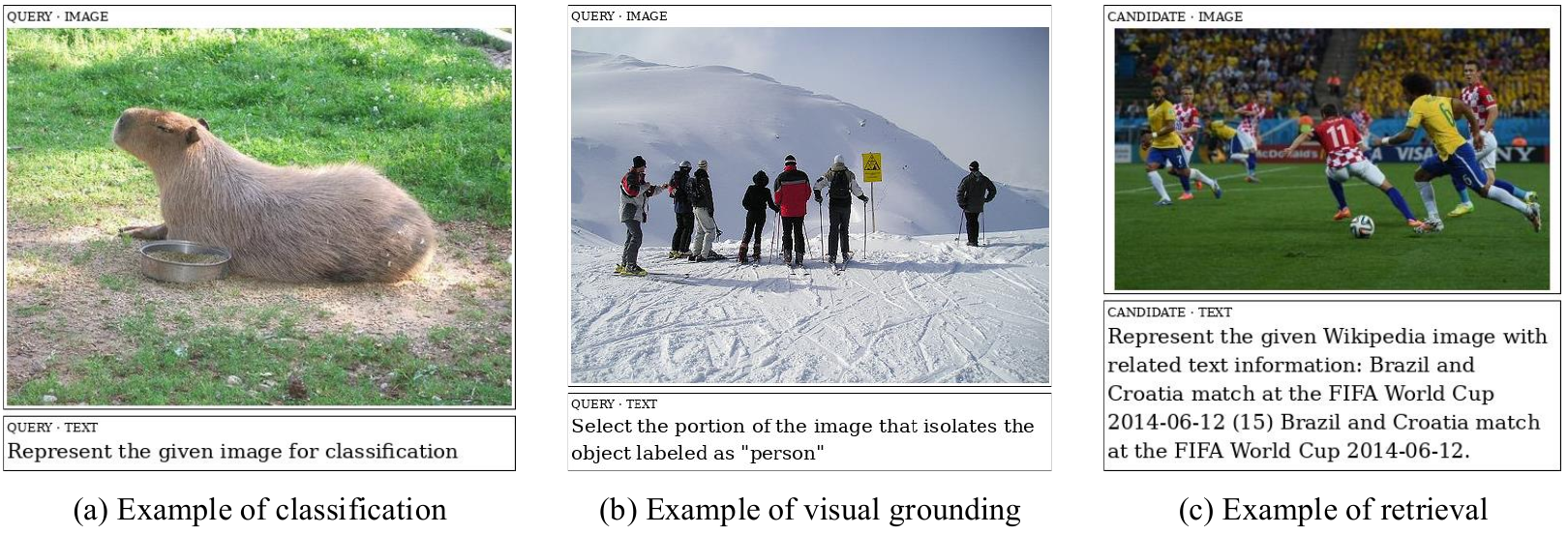}
    \caption{Single-stack rendering examples produced under the selected canonical rendering policy \(\psi^*\) across multimodal embedding tasks: (a) classification, (b) visual grounding, and (c) retrieval.}
    \label{fig:single_stack_examples}
    \vspace{0.5em}
    \includegraphics[width=0.9\linewidth]{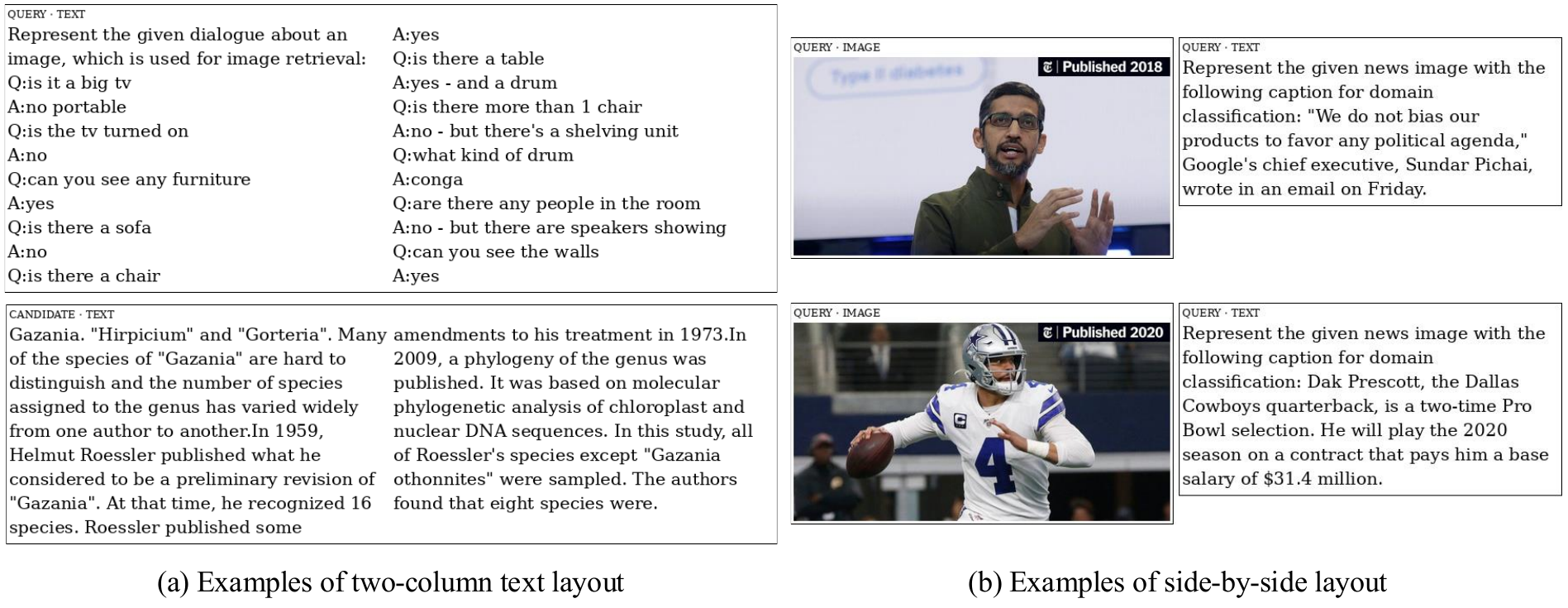}
    \caption{Composition-specific layout examples produced under the selected canonical rendering policy \(\psi^*\): (a) two-column text and (b) side-by-side text--image.}
    \label{fig:adaptive_layout_examples}
\end{figure*}

\paragraph{Adaptive Layout Router.}
Given the rendered element cards, the Adaptive Layout Router constructs the layout plans applicable to the input composition, as summarized in Table~\ref{tab:rendering_layouts}. Every composition admits an order-preserving vertical plan, denoted by \(\pi_{\mathrm{stack}}\). A single-text input additionally admits a two-column plan \(\pi_{\mathrm{text}}\), while an input containing one text element and one image element admits a side-by-side plan \(\pi_{\mathrm{mix}}\). Each composition-specific plan is considered feasible only when every column or lane retains sufficient content width after padding.

For a single-text input, \(\pi_{\mathrm{text}}\) wraps the text at the equal-column width and divides the resulting lines into two contiguous columns. Let \(W_{\mathrm{stack}}\) and \(H_{\mathrm{stack}}\) denote the dimensions of \(\pi_{\mathrm{stack}}\). The router selects \(\pi_{\mathrm{text}}\) when it is feasible and \(H_{\mathrm{stack}}/W_{\mathrm{stack}}\geq\eta_{\mathrm{text}}\), where \(\eta_{\mathrm{text}}\) specifies the height-to-width ratio at which the two-column layout is activated. For an input containing one text element and one image element, \(\pi_{\mathrm{mix}}\) places the two cards in equal-width lanes while preserving their input order. Let \(h_{\mathrm{T}}\) and \(h_{\mathrm{I}}\) denote the resulting card heights. The router selects \(\pi_{\mathrm{mix}}\) when it is feasible, \(\max(h_{\mathrm{T}},h_{\mathrm{I}})/ \min(h_{\mathrm{T}},h_{\mathrm{I}})\leq\eta_{\mathrm{mix}}\), and \(\ell(\pi_{\mathrm{mix}})<\ell(\pi_{\mathrm{stack}})\). Here, \(\eta_{\mathrm{mix}}\) limits the imbalance between the two card heights, and \(\ell(\pi)\) denotes the visual-token count produced by rasterizing plan \(\pi\). Inputs without an applicable composition-specific plan, or whose plan fails its routing criteria, use \(\pi_{\mathrm{stack}}\). Figures~\ref{fig:single_stack_examples} and
\ref{fig:adaptive_layout_examples} show representative renderings produced
under the selected canonical rendering policy \(\psi^*\), specified in
Section~\ref{app:selected_rendering_policy}.

Algorithm~\ref{alg:alg:visual_substrate_rendering} formalizes these routing rules and
rasterizes the selected plan into the visual substrate \(\psi(x)\).

\begin{algorithm*}[t]
    \caption{Visual Substrate Rendering}
    \label{alg:alg:visual_substrate_rendering}
    \begin{algorithmic}[1]
        \REQUIRE Retrieval unit \(x\), rendering policy \(\psi\)
        \ENSURE Common visual substrate \(\psi(x)\)
        \STATE Extract the non-empty text and image elements of \(x\)
        in input order as \(\mathcal U(x)\)
        \STATE Use the Element Renderer to render each element in
        \(\mathcal U(x)\) as a typed visual card under \(\psi\)
        \STATE Use the Adaptive Layout Router to stack the cards in input order,
        yielding \(\pi_{\mathrm{stack}}\) with dimensions
        \((W_{\mathrm{stack}},H_{\mathrm{stack}})\); set \(\pi\gets\pi_{\mathrm{stack}}\)
        \IF{\(\mathcal U(x)\) consists of a single text element}
            \STATE Wrap the text at the equal-column width and divide the
            resulting lines into two contiguous columns, yielding
            \(\pi_{\mathrm{text}}\)
            \IF{\(\pi_{\mathrm{text}}\) is feasible and
            \(H_{\mathrm{stack}}/W_{\mathrm{stack}}\geq
            \eta_{\mathrm{text}}\)}
                \STATE \(\pi\gets\pi_{\mathrm{text}}\)
            \ENDIF
        \ELSIF{\(\mathcal U(x)\) consists of one text element and one image element}
            \STATE Place the two cards in equal-width lanes in input order,
            yielding \(\pi_{\mathrm{mix}}\) with card heights
            \(h_{\mathrm{T}},h_{\mathrm{I}}\)
            \IF{\(\pi_{\mathrm{mix}}\) is feasible,
            \(\max(h_{\mathrm{T}},h_{\mathrm{I}})/
            \min(h_{\mathrm{T}},h_{\mathrm{I}})
            \leq\eta_{\mathrm{mix}}\), and
            \(\ell(\pi_{\mathrm{mix}})
            <\ell(\pi_{\mathrm{stack}})\)}
                \STATE \(\pi\gets\pi_{\mathrm{mix}}\)
            \ENDIF
        \ENDIF
        \RETURN the visual substrate \(\psi(x)\) assembled with the canvas extent
        and card placement determined by \(\pi\)
    \end{algorithmic}
\end{algorithm*}

\subsection{Retrieval-Guided Policy Search}
\label{app:rendering_policy_search}

Retrieval-guided policy search selects \(\psi^*\) under a visual-token budget. We construct a discrete policy space \(\Psi\) by varying the rendering and routing parameters while preserving the substrate-construction procedure in Section~\ref{app:rendering_construction}. After filtering geometrically invalid configurations, each remaining policy is evaluated on a common policy-search set \(\mathcal V\) with a frozen proxy retriever. The search returns the highest-scoring policy whose mean visual-token count does not exceed \(\bar{\ell}=480\), breaking score ties by lower visual-token cost.

\input{tables/appendix/search_space}

\paragraph{Candidate policies and feasibility.} We instantiate each candidate policy by assigning one value to every parameter in Table~\ref{tab:rendering_policy_space}. In particular, the routing thresholds allow different policies to select different layouts for the same input. Let \(p\), \(w_{\min}\), and \(w_{\max}\) denote the card padding, minimum substrate width, and maximum substrate width of a policy, respectively. We retain the policy-level feasible subset
\begin{equation}
    \Psi_{\mathrm{feas}}
    =
    \left\{\psi\in\Psi:\;2p<w_{\min}\leq w_{\max}\right\}.
\end{equation}
This screening is applied whenever a policy is initialized or generated, while the feasibility of a composition-specific layout remains input dependent as defined in Section~\ref{app:rendering_construction}.

\paragraph{Policy-search data and proxy scorer.} We construct \(\mathcal V\) from the training subsets of MoCa-CL-Pairs\footnote{\url{https://huggingface.co/datasets/moca-embed/MoCa-CL-Pairs}}. Using random seed 42, we sample 100 query--positive pairs from each subset. The sampled queries form the query set, and their distinct positive candidates form the retrieval pool for that subset. The resulting policy-search data come only from training subsets and do not include final evaluation examples.

Candidate policies are evaluated with a frozen MoCa-3B encoder. Queries and candidates are encoded from their rendered substrates using the role-specific instructions ``Represent this rendered query for retrieving a relevant document.'' and ``Represent this rendered document for retrieval.'' The scorer applies last-token pooling and \(\ell_2\) normalization and computes cosine similarities. We compute Precision@1 within each subset and macro-average it across subsets to obtain \(\mathcal M(\psi;\mathcal V)\). The same sampled queries and distinct candidates are used to measure \(\mathbb{E}_{x\sim\mathcal V}[\ell(\psi(x))]\), and policies exceeding \(\bar{\ell}=480\) are excluded from retrieval-guided search.

\begin{algorithm*}[t]
    \caption{Retrieval-Guided Policy Search}
    \label{alg:rendering_policy_search}
    \begin{algorithmic}[1]
        \REQUIRE Policy space \(\Psi\), policy-search set \(\mathcal V\),
        frozen proxy retriever; settings \(\bar{\ell}\), \(n_{\mathrm{pop}}\),
        \(T\), \(n_{\mathrm{elite}}\), \(\rho\), \(s\)
        \ENSURE Canonical rendering policy \(\psi^*\)
        \STATE Construct
        \(\Psi_{\mathrm{feas}}=\{\psi\in\Psi:2p<w_{\min}\leq w_{\max}\}\)
        \STATE Initialize all stochastic operations with seed \(s\)
        \STATE Draw \(n_{\mathrm{pop}}\) distinct policies from
        \(\Psi_{\mathrm{feas}}\) to form \(\mathcal P\); set \(\mathcal H\gets\varnothing\)
        \FOR{\(g=1,\ldots,T\)}
            \FORALL{\(\psi\in\mathcal P\setminus\mathcal H\)}
                \STATE Render \(\mathcal V\) with \(\psi\) and compute
                \(\mathbb{E}_{x\sim\mathcal V}[\ell(\psi(x))]\)
                \IF{\(\mathbb{E}_{x\sim\mathcal V}[\ell(\psi(x))]\leq\bar{\ell}\)}
                    \STATE Evaluate \(\psi\) with the frozen proxy retriever and set
                    \(\mathcal M(\psi;\mathcal V)\) to its macro-averaged Precision@1
                \ELSE
                    \STATE \(\mathcal M(\psi;\mathcal V)\gets-\infty\)
                \ENDIF
                \STATE Add \(\psi\) to the evaluated set \(\mathcal H\)
            \ENDFOR
            \IF{\(g<T\)}
                \STATE Let \(\mathcal E\) be the \(n_{\mathrm{elite}}\)
                budget-feasible policies in \(\mathcal P\) ranked by
                \(\mathcal M(\cdot;\mathcal V)\)
                \STATE Initialize the next population
                \(\mathcal P'\gets\mathcal E\)
                \WHILE{\(|\mathcal P'|<n_{\mathrm{pop}}\) and an unseen
                offspring is available}
                    \IF{two policies are available and a probability-\(\rho\)
                    crossover is selected}
                        \STATE Choose two policies from \(\mathcal E\) and
                        independently copy each parameter from either policy to
                        form \(\psi_{\mathrm{new}}\)
                    \ELSE
                        \STATE Mutate one parameter of a policy from \(\mathcal E\)
                        to form \(\psi_{\mathrm{new}}\)
                    \ENDIF
                    \IF{\(\psi_{\mathrm{new}}\in\Psi_{\mathrm{feas}}\) and has not
                    been evaluated or added}
                        \STATE Add \(\psi_{\mathrm{new}}\) to \(\mathcal P'\)
                    \ENDIF
                \ENDWHILE
                \STATE Fill any remaining slots in \(\mathcal P'\) with
                distinct unevaluated policies from \(\Psi_{\mathrm{feas}}\)
                \STATE \(\mathcal P\gets\mathcal P'\)
            \ENDIF
        \ENDFOR
        \RETURN the budget-feasible policy in \(\mathcal H\) with the largest
        \(\mathcal M(\psi;\mathcal V)\), breaking ties by lower
        \(\mathbb{E}_{x\sim\mathcal V}[\ell(\psi(x))]\)
    \end{algorithmic}
\end{algorithm*}

\paragraph{Evolutionary search.} We solve the constrained policy-selection problem with the elitist procedure in Algorithm~\ref{alg:rendering_policy_search}. We use a visual-token budget \(\bar{\ell}=480\), population size \(n_{\mathrm{pop}}=24\), elite count \(n_{\mathrm{elite}}=6\), and crossover rate \(\rho=0.5\). At each round, the highest-scoring budget-feasible policies seed the next population. Crossover independently inherits each parameter from one of two elite policies. Mutation changes a numerical parameter to an adjacent candidate value or changes a categorical parameter to a different value in its candidate set. Remaining population slots are filled with distinct unevaluated policies from \(\Psi_{\mathrm{feas}}\). After \(T=8\) rounds, the search returns the budget-feasible policy with the largest \(\mathcal M(\psi;\mathcal V)\), breaking ties by lower mean visual-token cost. All stochastic operations use random seed \(s=42\).

\begin{figure*}[t]
    \centering
    \begin{minipage}[t]{0.34\linewidth}
        \centering
        \includegraphics[width=\linewidth]{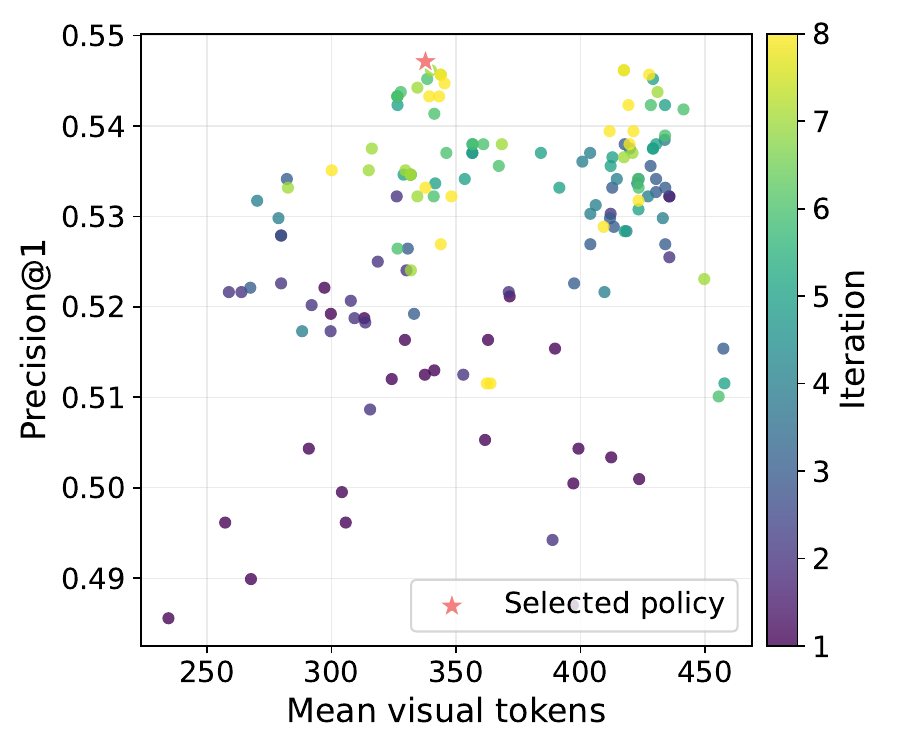}
        \small (a) Overall distribution
    \end{minipage}\hfill
    \begin{minipage}[t]{0.34\linewidth}
        \centering
        \includegraphics[width=\linewidth]{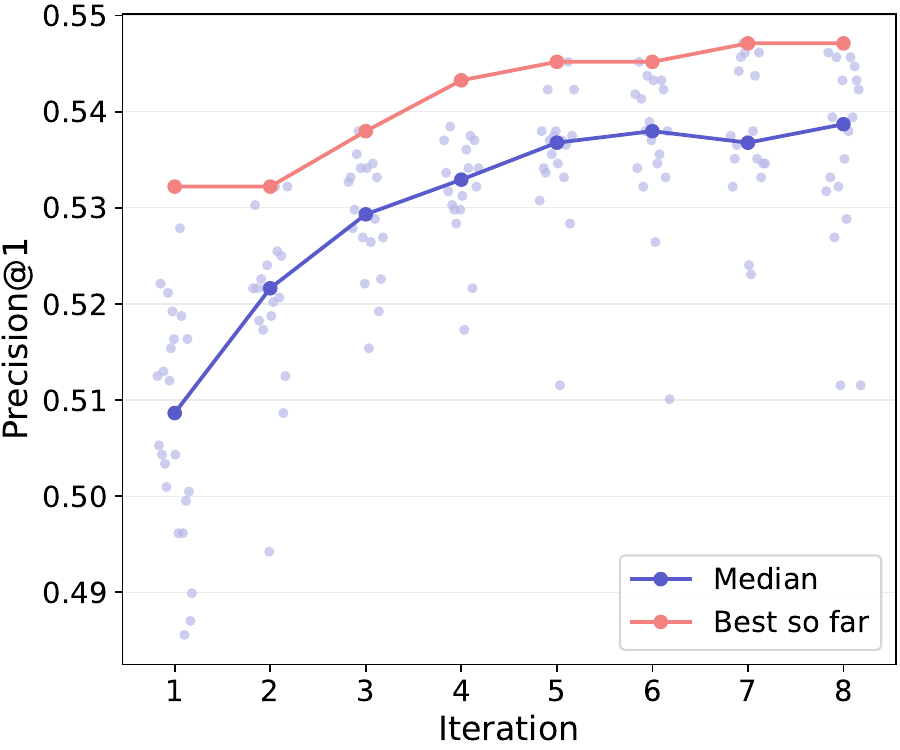}
        \small (b) Generation evolution
    \end{minipage}\hfill
    \begin{minipage}[t]{0.3\linewidth}
        \centering
        \includegraphics[width=\linewidth]{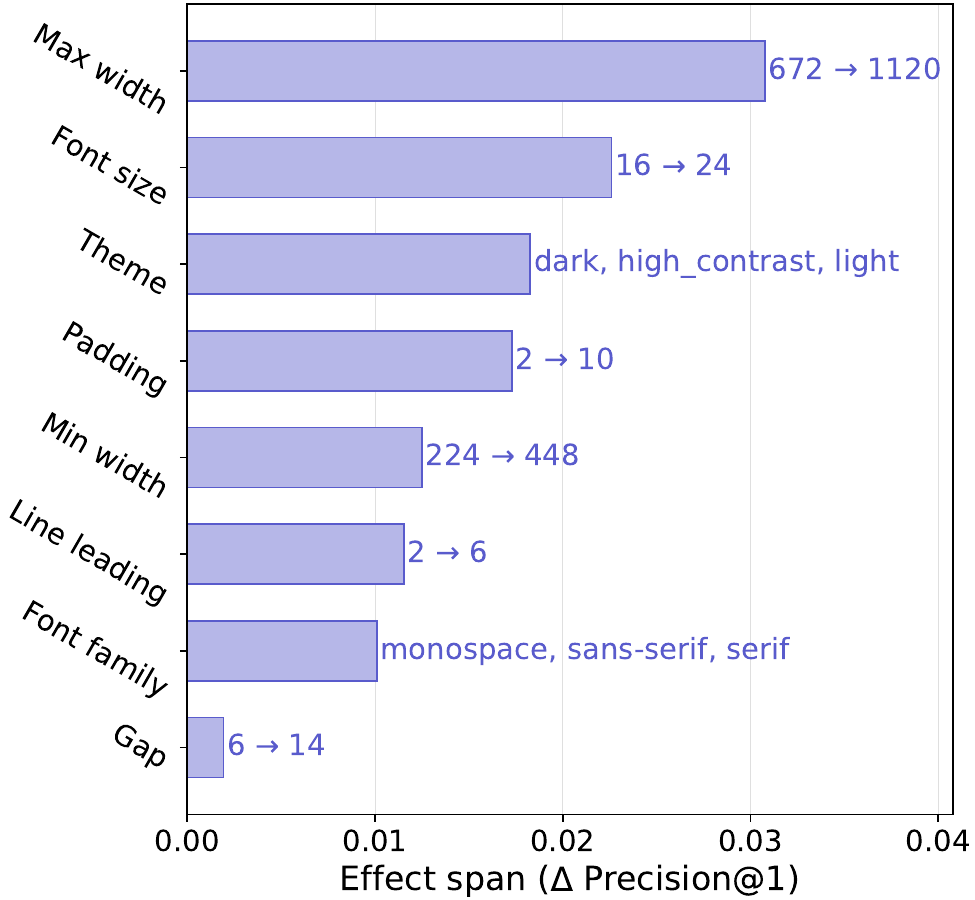}
        \small (c) One-factor sensitivity
    \end{minipage}
    \caption{Rendering-policy search results on \(\mathcal V\). (a) Candidate policies by macro-averaged Precision@1 and mean visual-token count, with color indicating the search iteration and the star marking the selected policy. (b) Evolution of the score distribution, median, and best-so-far policy across search iterations. (c) Sensitivity of the eight most influential factors around the selected policy, measured by the macro-averaged Precision@1 range obtained when varying each factor while holding the others fixed.}
    \label{fig:policy_search_results}
\end{figure*}

\subsection{Canonical Rendering Policy}
\label{app:selected_rendering_policy}

Figure~\ref{fig:policy_search_results}(a) shows the performance and visual-token cost of the evaluated candidate policies on \(\mathcal V\). Their macro-averaged Precision@1 ranges from approximately 0.485 to 0.547, while their mean visual-token counts range from approximately 230 to 460. As the search proceeds, the candidates progressively concentrate in the high-performance region. Among the budget-feasible evaluated candidates, the canonical rendering policy \(\psi^*\) attains the highest macro-averaged Precision@1, with ties resolved by lower visual-token cost. Defined by the values in Table~\ref{tab:rendering_policy_space}, \(\psi^*\) uses 338 mean visual tokens and achieves a macro-averaged Precision@1 of 0.547. This selection shows that strong retrieval performance does not require the largest rendering budget.

Figure~\ref{fig:policy_search_results}(b) traces performance across search generations. Over the first five generations, the median macro-averaged Precision@1 rises from approximately 0.509 to 0.537, while the best-so-far score increases from approximately 0.532 to 0.546. Subsequent improvements are small, indicating that the search has largely converged.

Figure~\ref{fig:policy_search_results}(c) presents a one-factor sensitivity analysis around \(\psi^*\). We vary one rendering parameter across its search range while holding all remaining parameters at their selected values, and measure the resulting range of macro-averaged Precision@1. The figure reports the eight most influential factors, while those with smaller observed ranges are omitted for clarity. Maximum width has the largest effect, with a score range of approximately 0.031, followed by body font size, color theme, and card padding. Minimum width, line leading, and font family have intermediate effects, while card gap has the smallest effect among the eight factors shown. These ranges measure sensitivity within the search space and do not imply that increasing a parameter monotonically improves performance.

\input{tables/appendix/training_configuration.tex}

\section{Additional Experimental Details}
\label{app:experimental_details}

\subsection{Training and Model Configuration}
\label{app:training_configuration}

\paragraph{Training data.} We train RenderEmbed on MoCa-CL-Pairs~\citep{chen2026moca}, which combines MMEB-train~\citep{jiang2024vlm2vec} and ViDoRe-train~\citep{faysse2025colpali} in a single training corpus. All retrieval units are rendered before encoding, following the input construction described in Section~3 of the main paper.

\paragraph{Model and optimization.} RenderEmbed uses Qwen2.5-VL-3B as its backbone. We train the role-specific query and candidate anchor banks, the cross-layer fusion module, and LoRA adapters inserted into every LLM layer, while keeping all remaining parameters frozen. LoRA uses rank 32 and scaling factor \(\alpha=32\). Both anchor banks use \(M=16\) anchors, and the fusion module combines anchor states from layers \(\mathcal I_{\mathrm{fuse}}=\{18,24,30,36\}\) into \(H=2048\)-dimensional retrieval vectors. We optimize the model with Adam using \(\beta_1=0.9\), \(\beta_2=0.999\), a learning rate of \(1\times10^{-4}\), and 100 warmup steps. Training runs for one epoch with a global batch size of 128, temperature \(\tau=0.02\), rendering-consistency weight \(\lambda_{\mathrm{rc}}=0.1\), and random seed 42. Table~\ref{tab:training_configuration} summarizes these settings.

\input{tables/appendix/backbones_baselines}

\paragraph{Baselines.} Table~\ref{app:backbones-baselines} lists the public checkpoints or implementations used for the baselines that we reproduce. All remaining baseline results reported in Tables~1 and~2 of the main paper are taken from VLM2Vec~\citep{jiang2024vlm2vec}, MetaEmbed~\citep{xiao2026metaembed}, and MAGIC~\citep{yuan2026magic}.

\input{tables/appendix/encoding_training_costs}

\subsection{Intervention Construction}
\label{app:intervention_details}

\paragraph{Rendering-controlled semantic negatives.} For composed image--text candidates whose relevance depends on both components, we retain one component of the positive and replace the other to construct partially correct but mismatched pairs. These negatives encourage the model to distinguish a fully relevant composition from a match to an individual component.

For each training triple \((q_i,d_i^+,d_i^-)\), we scan the elements of \(d_i^+\) in their input order and select the first element whose same-type counterpart in \(d_i^-\) has different content. Since each candidate contains at most one text element and one image element, element type uniquely determines the corresponding replacement. We construct the rendering-controlled negative \(\hat d_i^-\) by replacing only the selected content while preserving all untouched elements and the canvas, card geometry, column structure, and typography determined by the canonical rendering policy \(\psi^*\). Replacement text is limited to the length of the original text and clipped within its content region. Replacement images are fitted proportionally within the same region without cropping or upscaling. If no same-type element with different content is available, we use the original negative \(d_i^-\) rendered with \(\psi^*\).

\paragraph{Manual validation of semantic negatives.}
We sample 20 examples from each of the 26 training subsets and assess the validity of the resulting 520 constructed negatives. Two annotators judge whether each candidate is a valid negative for its query, reporting validity rates of 99\% and 98.8\%, respectively. We inspect the six examples flagged by either annotator. Three from \texttt{tevatron\_colpali} have identical positive and negative images stored under different filenames. Three from \texttt{MSCOCO\_i2t} contain negative captions that closely match the positive descriptions and are judged false negatives. All six observed false negatives originate from the source annotations.

\paragraph{Alternative rendering policies.} We construct the alternative rendering-policy pool by randomly sampling ten distinct non-canonical policies from the budget-feasible policies whose retrieval performance on the policy-search queries exceeds the mean performance of all policies evaluated during the search. The selected policies vary in visual style, layout geometry, and routing rules. At each training step, we sample one policy \(\psi'\) for each local batch and apply it to every positive candidate \(d_j^+\) in that batch, producing \(\psi'(d_j^+)\). The sampled policy preserves the content and order of all elements, while each query continues to use the canonical rendering \(\psi^*(q_i)\).

\subsection{Efficiency Analysis}
\label{app:efficiency_details}

\paragraph{Retrieval Efficiency.}
To isolate retrieval-stage efficiency, we construct a shared BF16 workload with 1,000 queries and 100,000 candidates. For each model, \(M_q\) and \(M_d\) denote the average numbers of query and candidate vectors, respectively, \(H\) denotes the embedding dimension, and \(\mathcal C\) denotes the candidate collection. Since each BF16 value occupies two bytes, the candidate-side index size is \(2|\mathcal C|M_dH\) bytes. Counting a multiply-add as two FLOPs, scoring one query against the complete collection requires \(2|\mathcal C|M_qM_dH\) FLOPs. In Table~5 of the main paper, average vector counts are rounded for display, while FLOPs use the unrounded averages. For late-interaction scoring latency, we measure execution on an NVIDIA H20 using query and candidate batch sizes of 1,024. We report latency per query, averaged over the 1,000-query workload and ten timed runs after three warm-up runs. The timing includes pairwise dot products and MaxSim reduction while excluding encoding overhead.

\paragraph{Encoding and Training Overhead.}
We compare with NativeEmbed to quantify the overhead of visual rendering and its training objectives under the same backbone and anchor/fusion architecture. We additionally disable the alternative rendering view to measure the cost of rendering-consistency training. Table~\ref{tab:encoding_training_costs} reports measurements on a single NVIDIA H20 with BF16, using 64 query--candidate pairs from each of 26 subsets (1,664 pairs in total). Encoding uses batch size 1 and measures latency from raw input to embedding, including preprocessing and rendering. Token counts include padding and anchors. Training uses batch size 4, with 10 warm-up steps followed by 20 timed steps per run. Alternative rendering views are used only during training, so the two RenderEmbed settings share the encoding measurements.

Visual rendering increases average input token counts by 17.8\% for queries and 68.1\% for candidates. Query encoding takes an additional 29.34 ms (21.4\%), while candidate encoding increases by 38.65 ms (40.2\%). Rendering itself accounts for 25.33 ms per query and 13.73 ms per candidate, already included in the encoding latency. The query overhead affects online retrieval, whereas candidate encoding is performed offline during index construction or updates.

Training throughput decreases from 2.381 pairs/s for NativeEmbed to 1.981 pairs/s for RenderEmbed without alternative views, and to 1.547 pairs/s with the full training objective. For a fixed number of training pairs, these rates imply approximately 20.2\% and 53.9\% more training time than NativeEmbed, respectively. The additional view increases training cost without changing the inference procedure.

\begin{figure*}[t]
    \centering
    \begin{minipage}[t]{0.3\textwidth}
        \centering
        \includegraphics[width=\linewidth]
        {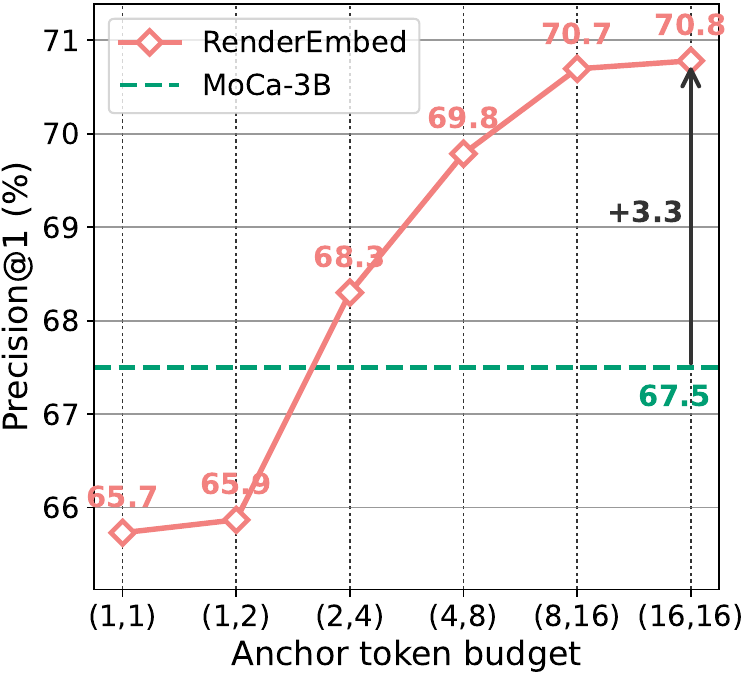}
        \small (a) Candidate-heavy: \(M_q \leq M_d\)
    \end{minipage}\hfill
    \begin{minipage}[t]{0.3\textwidth}
        \centering
        \includegraphics[width=\linewidth]
        {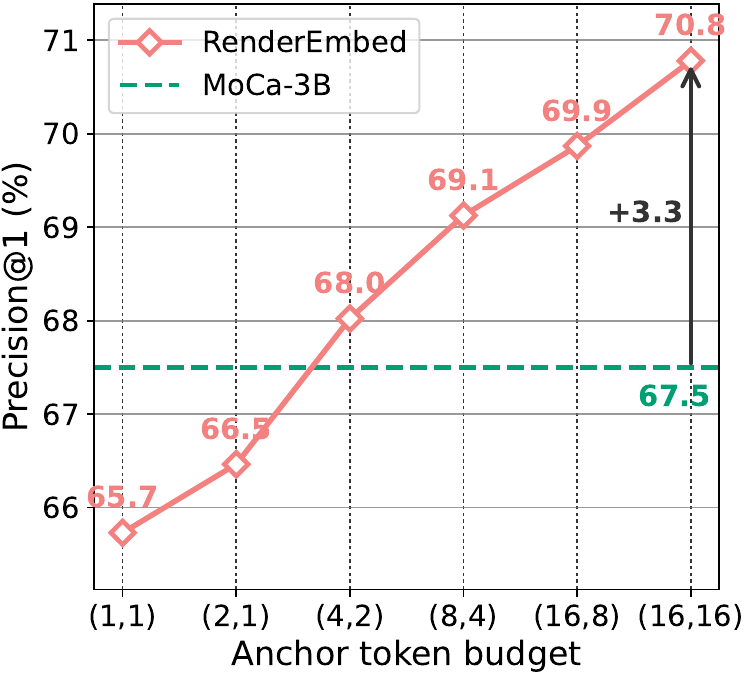}
        \small (b) Query-heavy: \(M_q \geq M_d\)
    \end{minipage}\hfill
    \begin{minipage}[t]{0.3\textwidth}
        \centering
        \includegraphics[width=\linewidth]
        {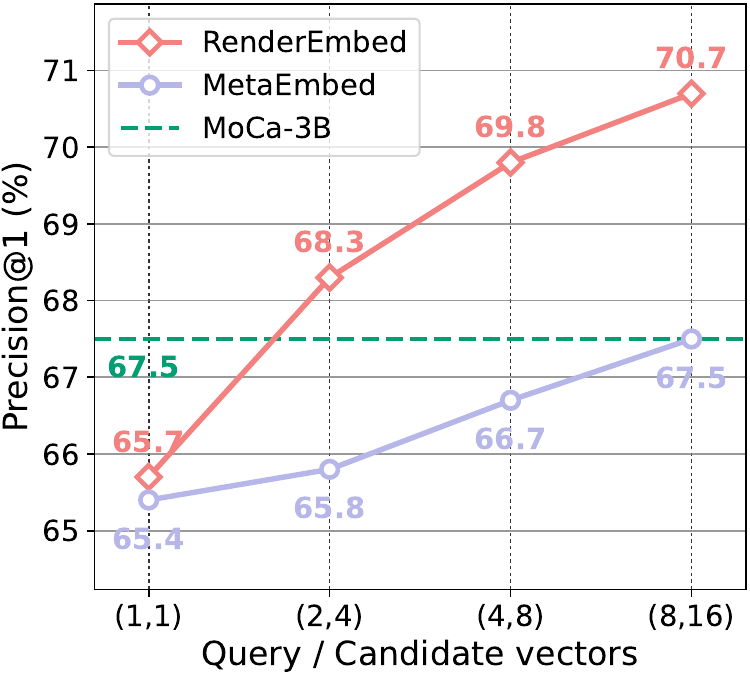}
        \small (c) Test-time budget comparison
    \end{minipage}
    \caption{Test-time vector budgets on MMEB. Each pair \((M_q,M_d)\) denotes the numbers of retained query and candidate vectors. (a) Candidate-heavy allocations, \(M_q \leq M_d\). (b) Query-heavy allocations, \(M_q \geq M_d\). Corresponding settings in (a) and (b) keep \(M_qM_d\) fixed, matching scoring FLOPs. (c) RenderEmbed and MetaEmbed-3B at identical query and candidate vector budgets. MoCa-3B provides an accuracy reference in all panels.}
    \label{fig:anchor_budget_asymmetry}
\end{figure*}

\section{Additional Experimental Results}
\label{app:additional_results}

\subsection{Comparison with Rendering Augmentation}
\label{app:rendering_augmentation}

We compare rendering consistency with ordinary rendering augmentation using the same 25\% training subset and experimental setup as Table~3 of the main paper. The augmentation baseline retains the model architecture and rendering policies, replacing margin consistency with direct contrastive supervision on alternative candidate views. Pairing canonical queries with alternatively rendered candidates gives $\widetilde{s}_{ij}=s\bigl(\psi^*(q_i),\psi'(d_j^+)\bigr)$, with losses
\begin{equation}
\begin{aligned}
\mathcal{L}_{\mathrm{alt\text{-}ret}}
&=-\frac{1}{B}\sum_{i=1}^{B}
\log\frac{\exp(\widetilde{s}_{ii}/\tau)}
{\sum_{j=1}^{B}\exp(\widetilde{s}_{ij}/\tau)},\\
\mathcal{L}_{\mathrm{aug}}
&=\mathcal{L}_{\mathrm{ret}}+0.1\,\mathcal{L}_{\mathrm{alt\text{-}ret}}.
\end{aligned}
\end{equation}
The alternative-view loss uses only in-batch negatives, while the original-view $\mathcal{L}_{\mathrm{ret}}$ retains the rendering-controlled semantic negatives in Eq.~(7) of the main paper. We evaluate canonical retrieval accuracy and its change under ten non-canonical rendering policies, following Section~4.4 of the main paper.

\input{tables/appendix/rendering_augmentation}

Table~\ref{tab:rendering_augmentation} shows that both augmentation and margin consistency reduce mean degradation from 0.38 to 0.12 points relative to removing the rendering intervention. At the same mean degradation, margin consistency achieves 0.7 points higher canonical accuracy, retaining retrieval quality alongside rendering robustness.

\subsection{Asymmetric Test-Time Anchor Budgets}
\label{app:asymmetric_anchor_budget}

We extend the symmetric-budget analysis to candidate-heavy allocations in Figure~\ref{fig:anchor_budget_asymmetry}(a) and query-heavy allocations in Figure~\ref{fig:anchor_budget_asymmetry}(b). At the smallest asymmetric budget, expanding the query representation is more beneficial. Once the query side retains several anchors, allocating additional anchors to candidates consistently produces stronger retrieval at matched computational cost. In particular, preserving the full candidate budget while reducing the query budget nearly maintains full-budget performance, whereas the converse allocation incurs a larger degradation. This pattern suggests that candidate coverage becomes the primary bottleneck after the query is represented by a sufficient number of anchors, since each query vector benefits from a richer set of candidate evidence under MaxSim. The two allocation directions therefore expose a practical trade-off: candidate-heavy settings favor retrieval quality, whereas query-heavy settings reduce candidate-index storage.

Figure~\ref{fig:anchor_budget_asymmetry}(c) compares RenderEmbed with MetaEmbed-3B at matched vector budgets, using MetaEmbed results from Table~5 of \citet{xiao2026metaembed}. Both models use 2,048-dimensional embeddings, so matching their vector budgets also matches candidate embedding storage and scoring FLOPs. Under these equal costs, RenderEmbed achieves 0.3--3.2 points higher Overall Precision@1 across all four budgets. MoCa-3B provides an accuracy reference for budget selection. RenderEmbed exceeds its score of 67.5 at \((2,4)\), reaching 68.3, while MetaEmbed matches it at \((8,16)\). The same RenderEmbed model reaches 70.7 at \((8,16)\) without retraining, offering flexible test-time control over accuracy, index storage, and scoring cost through query and candidate budget selection.

\input{tables/appendix/single_vector_transfer}

\subsection{Visual Rendering with VLM2Vec Architecture}
\label{app:single_vector_transfer}

To evaluate the effectiveness of visual rendering across architectures, we choose VLM2Vec~\citep{jiang2024vlm2vec}, an established single-vector multimodal retrieval architecture. We compare native and rendered multimodal inputs using Qwen2.5-VL-3B and the same 25\% training subset as Table~3 of the main paper. The rendered-input model uses Visual Substrate Rendering with its associated learning strategy, Interventional Rendering Learning, while retaining VLM2Vec's single-vector encoding.

As illustrated in Table~\ref{tab:single_vector_transfer}, visual rendering remains effective in this architecture: it improves all four meta-tasks and raises Overall Precision@1 from 64.7 to 67.3. The gain is larger on OOD tasks than on IND tasks, at 4.3 versus 1.3 points, extending the generalization benefit of unified visual inputs to single-vector architecture.

\input{tables/appendix/mmeb_v2}

\subsection{Evaluation on MMEB-V2}
\label{app:mmeb_v2}

MMEB-V2~\citep{meng2026vlmvecv} extends MMEB with 18 video tasks and 24 visual-document tasks. We evaluate RenderEmbed on all 42 additional tasks using the same model weights as in Table~1 of the main paper. For videos, Visual Substrate Rendering combines the keyframes sampled by the official evaluation code into a single image. Table~\ref{tab:mmeb_v2} reports Hit@1 for video tasks and NDCG@5 for visual-document tasks, alongside the baseline results from Table~2 of \citet{meng2026vlmvecv}.

RenderEmbed leads every task group. Its average video score reaches 47.5, exceeding the strongest reported baseline by 8.9 points. Rendering keyframes into a single image allows the trained model to process video inputs through its existing visual encoding pathway, without additional video training. On visual documents, RenderEmbed achieves an average score of 81.5, a 6.3-point gain over the strongest baseline. The largest advantage occurs on the OOD group, where the score increases from the best baseline's 44.4 to 70.5.

\subsection{Detailed Results on MMEB}

We present the detailed results of RenderEmbed and some representative baseline models on the MMEB benchmark in Table~\ref{tab:detail_mmeb}, covering 36 tasks across four categories: classification, VQA, retrieval, and visual grounding. 

\input{tables/appendix/grounding_resolution}

\begin{figure}[t]
    \centering
    \includegraphics[width=0.85\linewidth]{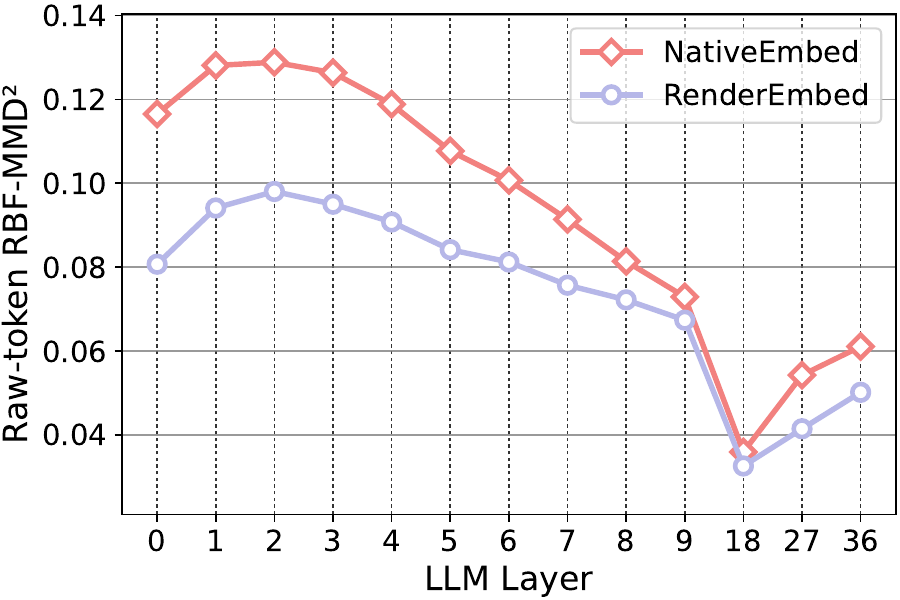}
    \caption{Layer-wise RBF-MMD$^2$ between caption- and image-derived content-token distributions on 1,000 MSCOCO pairs. Lower values indicate closer distributions. Layer 0 is the LLM input.}
    \label{fig:modality_differences}
\end{figure}

\section{Further Analysis}
\label{app:further_analysis}

\subsection{Modality Differences Across Layers}
\label{app:modality_differences}

Existing VLMs map text and images into a shared LLM input space through modality-specific input pathways. To assess whether unifying input formats before encoding is still necessary, we examine modality differences inside the shared LLM of NativeEmbed and RenderEmbed. Using their trained checkpoints, we separately encode the captions and images from the same 1,000 MSCOCO text-to-image evaluation pairs under the candidate role.

After a full forward pass, we uniformly sample eight content-token positions per input, excluding instructions, anchors, and other special tokens. We extract the raw vectors at these same positions across layers without normalization. At each layer, we compute squared maximum mean discrepancy with an RBF kernel (RBF-MMD$^2$) between the resulting 8,000 caption-derived and 8,000 image-derived vectors, including diagonal terms. All comparisons share a fixed bandwidth, estimated from pooled features across models and layers. Layer 0 denotes the LLM input, and layer 36 is the final block output.

Interestingly, Figure~\ref{fig:modality_differences} shows that MMD$^2$ decreases toward the middle layers and rises again in deeper layers, resembling the pattern reported by \citet{huang2025deciphering} for unnormalized cross-modal distances. This increase may partly reflect larger token magnitudes in deeper layers.

RenderEmbed exhibits lower MMD$^2$ at all 13 measured positions, with reductions of 30.7\% at the LLM input and 17.9\% at layer 36. From the input to layer 9, NativeEmbed's MMD$^2$ decreases more, narrowing the difference between the two models' MMD$^2$ values by 84.5\%. RenderEmbed thus starts with closer modality-dependent token distributions and maintains lower distribution discrepancy across layers, supporting our motivation for unifying inputs before the shared LLM.

\subsection{Input Unification in Text Retrieval}
\label{app:input_configurations}

To isolate the role of input unification, we conduct a text-only retrieval experiment in which the same underlying content is presented as native text or rendered images. We use the WebQA text-to-text subset of M-BEIR~\citep{wei2024uniir} and compare four input configurations: TT, II, TI, and IT. T denotes native text, and I denotes the same text rendered as an image, with the query format listed first. All four configurations use Qwen2.5-VL-3B backbone with the same training setup. Each model is trained with standard contrastive learning without intervention losses for one epoch on 16,097 queries, using a batch size of 128 and 10 warmup steps. Evaluation covers 2,455 queries and 544,457 text candidates.

\begin{figure}[t]
    \centering
    \includegraphics[width=\linewidth]{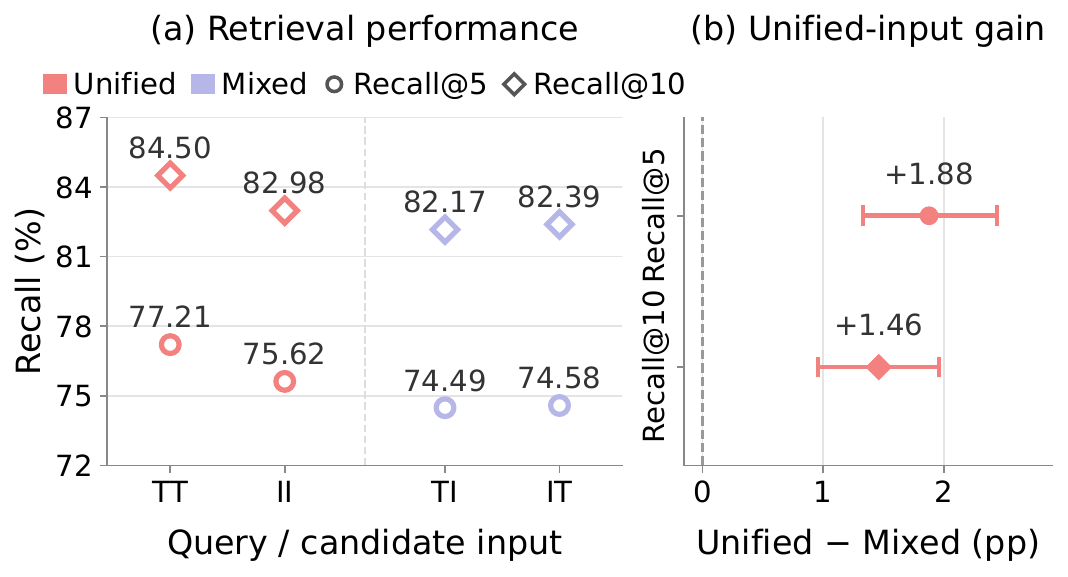}
    \caption{Input configurations on text-to-text retrieval. T denotes native text, and I denotes the same text rendered as an image, with the query format listed first. (a) Recall@5 and Recall@10 for unified (TT, II) and mixed (TI, IT) configurations. (b) Difference between their group means in percentage points. Error bars show 95\% paired query-bootstrap intervals from 10,000 resamples of the evaluation queries, keeping the four trained models fixed.}
    \label{fig:input_configurations}
\end{figure}

Figure~\ref{fig:input_configurations}(a) shows that both unified configurations outperform both mixed configurations on both metrics. Averaging TT and II against TI and IT yields gains of 1.88 percentage points in Recall@5 and 1.46 in Recall@10, with both confidence intervals above zero in Figure~\ref{fig:input_configurations}(b). TT exceeds II by 1.59 and 1.52 points, respectively. This may reflect the backbone's language pretraining, which favors processing native text. The preferred candidate format follows the query format: TT exceeds TI for native-text queries, while II exceeds IT for rendered queries. These results support input unification in both textual and visual forms, with native text performing best on this task.

Although native text performs better on this text-only task, it cannot directly accommodate natural images, visual documents, or text--image compositions. Instead, we extend input unification to multimodal retrieval through visual rendering. A visual canvas directly accommodates images and readable text, while converting images into textual descriptions can omit retrieval-relevant visual details. Together with the cross-configuration transfer results in Section~4.4 of the main paper, these findings support our interpretation that RenderEmbed's gains arise from unifying heterogeneous inputs through a shared visual encoding pathway. This common pathway allows supervision from different tasks to optimize the same retrieval mapping.

\subsection{Resolution Effects on Grounding}
\label{app:grounding_resolution}

To investigate RenderEmbed's weaker grounding performance in Table~1 of the main paper, we compare it with NativeEmbed using the same backbone across the four MMEB grounding subsets. Table~\ref{tab:grounding_resolution} shows an average of 83.2 for both models, with RenderEmbed's largest deficit on MSCOCO: 67.9 versus 74.5. Performance is similar on Visual7W-Pointing and higher on both RefCOCO subsets.

MSCOCO grounding requires retrieving an object crop from a full scene image and a target-category instruction. For 33.1\% of queries, the positive crop contains fewer than 4,096 pixels, and RenderEmbed achieves 32.6\% Precision@1 on this group. The renderer places these crops alongside instruction text without enlarging them. Text and whitespace can make the canvas satisfy the processor's minimum pixel budget while the object remains small. NativeEmbed instead resizes each candidate image independently, allowing small crops to reach the minimum image resolution. This difference suggests that insufficient local object resolution contributes to the MSCOCO degradation.

To test this explanation, we enlarge undersized candidate images before composing the canvas, following the native processor's resizing rule and minimum image area of 50,176 pixels. Applied at inference without retraining, this adjustment raises MSCOCO Precision@1 from 67.9 to 72.0 and the grounding average from 83.2 to 85.3. Preserving local object resolution during rendering thus alleviates the observed grounding degradation.

\input{tables/appendix/mmeb_details.tex}

%% file: tables/appendix/layout_template.tex
\begin{table}[t]
    \caption{Candidate layout templates. The schematics show the relative
    placement of element content. Exact dimensions are determined by the
    rendering policy and the input content.}
    \label{tab:rendering_layouts}
    \centering
    \setlength{\tabcolsep}{4pt}
    \renewcommand{\arraystretch}{1.15}
    \begin{tabular}{@{}
        >{\centering\arraybackslash}m{0.31\linewidth}
        >{\raggedright\arraybackslash}m{\dimexpr0.69\linewidth-2\tabcolsep\relax}
        @{}}
        \toprule
        Template & Applicability and routing \\
        \midrule
        \begin{tabular}[c]{@{}c@{}}
            Single stack\\[3pt]
            \fbox{\makebox[11mm][c]{\scriptsize Image}}\\[-1pt]
            \fbox{\makebox[11mm][c]{\scriptsize Text}}
        \end{tabular}
        & \textbf{Applicability:} Any supported composition.\par
        \textbf{Selection:} Baseline for every input. Retained when no applicable alternative
        satisfies its routing criteria. \\
        \addlinespace[4pt]
        \begin{tabular}[c]{@{}c@{}}
            Two-column text\\[3pt]
            \fbox{\makebox[18mm][c]{\scriptsize Text \(\mid\) Text}}
        \end{tabular}
        & \textbf{Applicability:} A single text element.\par
        \textbf{Selection:} Selected when both columns are feasible and the stacked layout is
        sufficiently tall. \\
        \addlinespace[4pt]
        \begin{tabular}[c]{@{}c@{}}
            Side-by-side\\[3pt]
            \fbox{\makebox[8mm][c]{\scriptsize\strut Image}}\hspace{1pt}%
            \fbox{\makebox[8mm][c]{\scriptsize\strut Text}}
        \end{tabular}
        & \textbf{Applicability:} One text and one image element.\par
        \textbf{Selection:} Selected when both lanes are feasible, their card heights are
        balanced, and the plan reduces the visual-token count. \\
        \bottomrule
    \end{tabular}
\end{table}

%% file: tables/appendix/search_space.tex
\begin{table*}[t]
    \caption{Rendering-policy search space and the selected canonical rendering policy.}
    \label{tab:rendering_policy_space}
    \centering
    \begin{tabular}{p{0.35\linewidth}p{0.35\linewidth}p{0.15\linewidth}}
        \toprule
        \textbf{Parameter} & \textbf{Candidate values} & \textbf{Selected} \\
        \midrule
        Minimum width (px)            & 224, 336, 400, 448            & 448 \\
        Maximum width (px)            & 672, 784, 896, 1008, 1120     & 896 \\
        Body font size (px)           & 16, 18, 20, 22, 24            & 22 \\
        Header font size (px)         & 10, 12, 14, 16                & 12 \\
        Line leading (px)             & 2, 4, 6                       & 4 \\
        Card padding (px)             & 2, 4, 6, 8, 10                & 4 \\
        Card gap (px)                 & 6, 8, 10, 14                  & 6 \\
        Font family                   & Sans serif, serif, monospace  & serif \\
        Color theme                   & Light, high contrast, dark    & high contrast \\
        Single-text aspect-ratio threshold \(\eta_{\mathrm{text}}\)
                                        & 0.02, 0.05, 0.1, 0.15, 0.18   & 0.15 \\
        Mixed-card height-ratio threshold \(\eta_{\mathrm{mix}}\)
                                        & 1.5, 2.0, 3.0, 4.0, 6.0       & 2.0 \\
        \bottomrule
    \end{tabular}
\end{table*}

%% file: tables/appendix/training_configuration.tex
\begin{table}[t]
\caption{Model and training configuration for RenderEmbed.}
\label{tab:training_configuration}
\centering
\resizebox{\linewidth}{!}{
\begin{tabular}{ll}
    \toprule
    \textbf{Configuration} & \textbf{Value} \\
    \midrule
    Anchors per query/candidate bank \(M\) & 16 \\
    Fusion layers \(\mathcal I_{\mathrm{fuse}}\) & \(\{18, 24, 30, 36\}\) \\
    Embedding dimension \(H\) & 2048 \\
    LoRA rank & 32 \\
    LoRA scaling alpha & 32 \\
    Adam \(\beta_1\) & 0.9 \\
    Adam \(\beta_2\) & 0.999 \\
    Learning rate & \(1 \times 10^{-4}\) \\
    Warmup steps & 100 \\
    Global batch size & 128 \\
    Training epochs & 1 \\
    Temperature \(\tau\) & 0.02 \\
    Rendering-consistency weight \(\lambda_{\mathrm{rc}}\) & 0.1 \\
    Random seed & 42 \\
    \bottomrule
\end{tabular}
}
\end{table}

%% file: tables/appendix/backbones_baselines.tex
\begin{table}[t]
\caption{Public sources for the baseline results reproduced in our experiments.}
\label{app:backbones-baselines}
\centering
\footnotesize
\begin{tabular}{@{}p{0.25\linewidth}p{0.7\linewidth}@{}}
\toprule
\textbf{Model} & \textbf{Public source} \\
\midrule
GME~\citep{zhang2024gme}                  & \url{https://huggingface.co/Alibaba-NLP/gme-Qwen2-VL-7B-Instruct} \\
UniME-V2-2B~\citep{gu2026unime}           & \url{https://huggingface.co/TianchengGu/UniME-V2-Qwen2VL-2B} \\
UniME-V2-7B~\citep{gu2026unime}           & \url{https://huggingface.co/TianchengGu/UniME-V2-Qwen2VL-7B} \\
ColQwen2.5~\citep{faysse2025colpali}      & \url{https://huggingface.co/vidore/colqwen2.5-v0.2} \\
Jina-v4~\citep{gunther2025jina}           & \url{https://huggingface.co/jinaai/jina-embeddings-v4} \\
AGC~\citep{qin2026multi}                  & \url{https://huggingface.co/hltcoe/AGC_qwen2.5-vl_colpali} \\
\bottomrule
\end{tabular}
\end{table}

%% file: tables/appendix/encoding_training_costs.tex
\begin{table*}[t]
\centering
\caption{Encoding and training costs. Input tokens are averaged; encoding latency and training throughput report mean $\pm$ standard deviation over five runs. Throughput counts each original query--candidate pair once, including when an alternative view is added.}
\label{tab:encoding_training_costs}
\resizebox{0.95\linewidth}{!}{
\begin{tabular}{lccccc}
\toprule
\multirow{2}{*}{\textbf{Method}} &
\multicolumn{2}{c}{\textbf{Input tokens} $\downarrow$} &
\multicolumn{2}{c}{\textbf{Encoding latency (ms)} $\downarrow$} &
\multirow{2}{*}{\textbf{Training throughput (pairs/s)} $\uparrow$} \\
\cmidrule(lr){2-3}\cmidrule(lr){4-5}
 & \textbf{Query} & \textbf{Candidate} & \textbf{Query} & \textbf{Candidate} & \\
\midrule
NativeEmbed & 385.5 & 167.6 & $136.97\pm1.77$ & $96.25\pm0.40$ & $2.381\pm0.003$ \\
RenderEmbed (w/o alternative view) & 454.2 & 281.7 & $166.31\pm0.41$ & $134.90\pm0.61$ & $1.981\pm0.005$ \\
RenderEmbed & 454.2 & 281.7 & $166.31\pm0.41$ & $134.90\pm0.61$ & $1.547\pm0.002$ \\
\bottomrule
\end{tabular}
}
\end{table*}

%% file: tables/appendix/rendering_augmentation.tex
\begin{table}[t]
\centering
\caption{Rendering consistency versus ordinary augmentation on MMEB. Overall is canonical Precision@1 (\%). $\Delta$P@1 is the change relative to each model's canonical score, reported as mean $\pm$ standard deviation across ten rendering policies.}
\label{tab:rendering_augmentation}
\resizebox{0.85\linewidth}{!}{
\begin{tabular}{lcc}
\toprule
\textbf{Method} & \textbf{Overall} & \textbf{$\Delta$P@1 (pp)} \\
\midrule
RenderEmbed & \textbf{68.6} & $-0.12\pm0.09$ \\
\textit{w/o Rendering Intervention} & 68.4 & $-0.38\pm0.32$ \\
\textit{w Rendering Augmentation} & 67.9 & $-0.12\pm0.11$ \\
\bottomrule
\end{tabular}
}
\end{table}

%% file: tables/appendix/single_vector_transfer.tex
\begin{table}[t]
\centering
\caption{Visual rendering on the single-vector VLM2Vec architecture. Precision@1 (\%) is reported on MMEB using the same task groups as Table~1 of the main paper. CLS, VQA, RET, and GND denote classification, visual question answering, retrieval, and grounding.}
\label{tab:single_vector_transfer}
\setlength{\tabcolsep}{4pt}
\resizebox{\linewidth}{!}{
\begin{tabular}{lccccccc}
\toprule
\multirow{2}{*}{\textbf{Input format}} &
\multicolumn{4}{c}{\textbf{Per Meta-Task Score}} & \multicolumn{3}{c}{\textbf{Average Score}} \\
\cmidrule(lr){2-5} \cmidrule(lr){6-8}
 & CLS & VQA & RET & GND & IND & OOD & Overall \\
\midrule
Native inputs & 59.2 & 61.6 & 66.4 & 81.0 & 69.0 & 59.3 & 64.7 \\
\rowcolor{gray!15} Rendered inputs & \textbf{62.6} & \textbf{65.3} & \textbf{67.5} & \textbf{83.8} & \textbf{70.3} & \textbf{63.6} & \textbf{67.3} \\
\bottomrule
\end{tabular}
}
\end{table}

%% file: tables/appendix/mmeb_v2.tex
\begin{table*}[t]
\centering
\caption{Results on the 42 tasks added in MMEB-V2. Baseline results and task groups follow Table~2 of \citet{meng2026vlmvecv}. CLS, QA, RET, and MRET denote classification, question answering, retrieval, and moment retrieval; VDR and VR denote ViDoRe and VisRAG. All scores are percentages.}
\label{tab:mmeb_v2}
\footnotesize
\resizebox{0.95\linewidth}{!}{
\begin{tabular}{lccccccccccc}
\toprule
\multirow{2}{*}{\textbf{Model}} & \multirow{2}{*}{\textbf{Size}} &
\multicolumn{5}{c}{\textbf{Video (Hit@1)}} & \multicolumn{5}{c}{\textbf{Visual Document (NDCG@5)}} \\
\cmidrule(lr){3-7} \cmidrule(lr){8-12}
 & & CLS & QA & RET & MRET & Overall & VDRv1 & VDRv2 & VR & OOD & Overall \\
\midrule
\multicolumn{2}{l}{\# Datasets} & 5 & 5 & 5 & 3 & 18 & 10 & 4 & 6 & 4 & 24 \\
\midrule
ColPali v1.3~\citep{faysse2025colpali} & 3B & 26.7 & 37.8 & 21.6 & 25.5 & 28.2 & 83.6 & 52.0 & 81.1 & 43.1 & 71.0 \\
GME~\citep{zhang2024gme} & 2B & 34.9 & 42.0 & 25.6 & 32.4 & 33.9 & 86.1 & 54.0 & 82.5 & 43.1 & 72.7 \\
GME~\citep{zhang2024gme} & 7B & 37.4 & 50.4 & 28.4 & 38.2 & 38.6 & 89.4 & 55.6 & 85.0 & 44.4 & 75.2 \\
LamRA-Qwen2~\citep{liu2025lamra} & 7B & 39.3 & 42.6 & 24.3 & 34.6 & 35.2 & 22.0 & 11.5 & 37.4 & 21.0 & 23.9 \\
LamRA-Qwen2.5~\citep{liu2025lamra} & 7B & 32.9 & 42.6 & 23.2 & 37.6 & 33.7 & 56.3 & 33.3 & 58.2 & 40.1 & 50.2 \\
VLM2Vec-Qwen2VL~\citep{jiang2024vlm2vec} & 2B & 33.4 & 30.5 & 20.6 & 33.0 & 29.0 & 49.8 & 13.5 & 51.8 & 33.5 & 41.6 \\
VLM2Vec-Qwen2VL~\citep{jiang2024vlm2vec} & 7B & 39.1 & 30.0 & 29.0 & 40.6 & 34.0 & 56.9 & 9.4 & 59.1 & 38.1 & 46.4 \\
VLM2Vec-V2~\citep{meng2026vlmvecv} & 2B & 39.3 & 34.3 & 28.8 & 38.5 & 34.9 & 75.5 & 44.9 & 79.4 & 39.4 & 65.4 \\
\midrule
\rowcolor{gray!15} RenderEmbed (Ours) & 3B & \textbf{45.7} & \textbf{54.6} & \textbf{43.3} & \textbf{45.6} & \textbf{47.5} & \textbf{90.2} & \textbf{59.8} & \textbf{88.6} & \textbf{70.5} & \textbf{81.5} \\
\bottomrule
\end{tabular}
}
\end{table*}

%% file: tables/appendix/grounding_resolution.tex
\begin{table*}[t]
\centering
\caption{Grounding Precision@1 (\%) on the four MMEB subsets. Adaptive enlargement resizes small candidate images before rendering at inference, without retraining. Avg. is the mean across the four subsets.}
\label{tab:grounding_resolution}
\resizebox{0.9\linewidth}{!}{
\begin{tabular}{lccccc}
\toprule
\textbf{Method} & \textbf{MSCOCO} & \textbf{Visual7W-Pointing} & \textbf{RefCOCO} & \textbf{RefCOCO-Matching} & \textbf{Avg.} \\
\midrule
NativeEmbed & \textbf{74.5} & 85.3 & 86.8 & 86.1 & 83.2 \\
RenderEmbed & 67.9 & 85.1 & 92.5 & \textbf{87.4} & 83.2 \\
\rowcolor{gray!15} RenderEmbed + adaptive enlargement & 72.0 & \textbf{87.9} & \textbf{94.0} & \textbf{87.4} & \textbf{85.3} \\
\bottomrule
\end{tabular}
}
\end{table*}

%% file: tables/appendix/mmeb_details.tex
\begin{table*}[ht]
\centering
\caption{Detailed performance on 36 MMEB tasks~\cite{jiang2024vlm2vec}. We show the results of our RenderEmbed and baseline models. Rows in yellow indicate metrics of an OOD task.}
\renewcommand{\arraystretch}{0.9}
\setlength{\tabcolsep}{2.5pt}
\resizebox{0.97\textwidth}{!}{
\begin{tabular}{lccccccccc}
\toprule
{Task} & {CLIP} & {SigLIP} & {BLIP2} & {VLM2Vec} & {GME} & {MMRet} & {UniME-V2} & {MoCa-3B} & {RenderEmbed} \\
\midrule
\multicolumn{10}{l}{{Classification (10 tasks)}} \\
ImageNet-1K                                        & 55.8 & 45.4 & 10.3 & 74.5 & 66.0 & 58.8 & 75.7 & 75.4 & 81.1 \\
N24News                                            & 34.7 & 13.9 & 36.0 & 80.3 & 52.0 & 71.3 & 43.3 & 80.9 & 79.6 \\
HatefulMemes                                       & 51.1 & 47.2 & 49.6 & 67.9 & 51.8 & 53.7 & 61.9 & 70.6 & 64.5 \\
VOC2007                                            & 50.7 & 64.3 & 52.1 & 91.5 & 79.0 & 85.0 & 46.1 & 87.0 & 85.7 \\
SUN397                                             & 43.4 & 39.6 & 34.5 & 75.8 & 71.3 & 70.0 & 63.9 & 74.8 & 78.4 \\
\rowcolor[HTML]{FFEB99} Place365                   & 28.5 & 20.0 & 21.5 & 44.0 & 40.3 & 43.0 & 34.7 & 38.8 & 44.4 \\
\rowcolor[HTML]{FFEB99} ImageNet-A                 & 25.5 & 42.6 & 3.2  & 43.6 & 40.8 & 36.1 & 48.2 & 39.7 & 42.3 \\
\rowcolor[HTML]{FFEB99} ImageNet-R                 & 75.6 & 75.0 & 39.7 & 79.8 & 83.1 & 71.6 & 85.1 & 75.4 & 82.8 \\
\rowcolor[HTML]{FFEB99} ObjectNet                  & 43.4 & 40.3 & 20.6 & 39.6 & 71.2 & 55.8 & 54.3 & 31.3 & 60.5 \\
\rowcolor[HTML]{FFEB99} Country-211                & 19.2 & 14.2 & 2.5  & 14.7 & 30.1 & 14.7 & 5.9  & 24.0 & 23.2 \\
\rowcolor[HTML]{E2F0D9}\textit{All Classification} & 42.8 & 40.3 & 27.0 & 61.2 & 58.6 & 56.0 & 51.9 & 59.8 & 64.3 \\
\midrule                                                                                                          
\multicolumn{10}{l}{{VQA (10 tasks)}}                                                                             \\
OK-VQA                                             & 7.5  & 2.4  & 8.7  & 69.0 & 33.1 & 73.3 & 57.5 & 40.0 & 70.7 \\
A-OKVQA                                            & 3.8  & 1.5  & 3.2  & 54.4 & 21.9 & 56.7 & 50.5 & 54.6 & 59.3 \\
DocVQA                                             & 4.0  & 4.2  & 2.6  & 52.0 & 44.2 & 78.5 & 88.1 & 93.0 & 95.3 \\
InfographicsVQA                                    & 4.6  & 2.7  & 2.0  & 30.7 & 22.4 & 39.3 & 56.5 & 67.7 & 73.4 \\
ChartQA                                            & 1.4  & 3.0  & 0.5  & 34.8 & 18.0 & 41.7 & 47.3 & 64.1 & 68.7 \\
Visual7W                                           & 4.0  & 1.2  & 1.3  & 49.8 & 23.6 & 49.5 & 55.2 & 61.6 & 58.5 \\
\rowcolor[HTML]{FFEB99} ScienceQA                  & 9.4  & 7.9  & 6.8  & 42.1 & 29.0 & 45.2 & 41.5 & 45.4 & 52.7 \\
\rowcolor[HTML]{FFEB99} VizWiz                     & 8.2  & 2.3  & 4.0  & 43.0 & 40.1 & 51.7 & 40.3 & 52.3 & 54.1 \\
\rowcolor[HTML]{FFEB99} GQA                        & 41.3 & 57.5 & 9.7  & 61.2 & 77.0 & 59.0 & 55.8 & 66.9 & 72.4 \\
\rowcolor[HTML]{FFEB99} TextVQA                    & 7.0  & 1.0  & 3.3  & 62.0 & 48.6 & 79.0 & 71.4 & 83.1 & 84.3 \\
\rowcolor[HTML]{E2F0D9}\textit{Avg. VQA}           & 9.1  & 8.4  & 4.2  & 49.9 & 35.8 & 57.4 & 56.4 & 62.9 & 68.9 \\
\midrule                                                                                                          
\multicolumn{10}{l}{{Retrieval (12 tasks)}}                                                                       \\
VisDial                                            & 30.7 & 21.5 & 18.0 & 80.9 & 65.9 & 83.0 & 33.2 & 80.5 & 81.0 \\
CIRR                                               & 12.6 & 15.1 & 9.8  & 49.9 & 60.0 & 61.4 & 57.4 & 55.7 & 65.2 \\
VisualNews\_t2i                                    & 78.9 & 51.0 & 48.1 & 75.4 & 81.6 & 74.2 & 69.2 & 74.4 & 74.3 \\
VisualNews\_i2t                                    & 79.6 & 52.4 & 13.5 & 80.0 & 84.2 & 78.1 & 79.9 & 77.8 & 78.7 \\
MSCOCO\_t2i                                        & 59.5 & 58.3 & 53.7 & 75.7 & 72.7 & 78.6 & 76.9 & 76.4 & 77.8 \\
MSCOCO\_i2t                                        & 57.7 & 55.0 & 20.3 & 73.1 & 59.8 & 72.4 & 70.6 & 72.6 & 74.9 \\
NIGHTS                                             & 60.4 & 62.9 & 56.5 & 65.5 & 68.4 & 68.3 & 65.4 & 67.4 & 69.0 \\
WebQA                                              & 67.5 & 58.1 & 55.4 & 87.6 & 91.3 & 90.2 & 89.4 & 90.6 & 89.9 \\
\rowcolor[HTML]{FFEB99} FashionIQ                  & 11.4 & 20.1 & 9.3  & 16.2 & 39.3 & 54.9 & 14.0 & 22.2 & 25.0 \\
\rowcolor[HTML]{FFEB99} Wiki-SS-NQ                 & 55.0 & 55.1 & 28.7 & 60.2 & 77.1 & 24.9 & 47.1 & 73.3 & 80.3 \\
\rowcolor[HTML]{FFEB99} OVEN                       & 41.1 & 56.0 & 39.5 & 56.5 & 79.0 & 87.5 & 65.4 & 75.9 & 75.3 \\
\rowcolor[HTML]{FFEB99} EDIS                       & 81.0 & 23.6 & 54.4 & 87.8 & 97.6 & 65.6 & 85.3 & 80.8 & 91.9 \\
\rowcolor[HTML]{E2F0D9}\textit{Avg. Retrieval}     & 53.0 & 31.6 & 33.9 & 67.4 & 73.1 & 69.9 & 62.8 & 70.6 & 73.6 \\
\midrule                                                                                                           
\multicolumn{10}{l}{{Visual Grounding (4 tasks)}}                                                                 \\
MSCOCO                                             & 33.8 & 46.4 & 28.9 & 80.6 & 48.0 & 76.8 & 57.9 & 80.2 & 67.9 \\
\rowcolor[HTML]{FFEB99} RefCOCO                    & 56.9 & 70.8 & 47.4 & 88.7 & 65.3 & 89.8 & 82.6 & 92.1 & 92.5 \\
\rowcolor[HTML]{FFEB99} RefCOCO-matching           & 61.3 & 50.8 & 59.5 & 84.0 & 82.3 & 90.6 & 88.8 & 92.8 & 87.4 \\
\rowcolor[HTML]{FFEB99} Visual7W-pointing          & 55.1 & 70.1 & 52.0 & 90.9 & 73.7 & 77.0 & 67.8 & 89.5 & 85.1 \\
\rowcolor[HTML]{E2F0D9}\textit{Avg. Grounding}     & 51.8 & 59.5 & 47.0 & 86.1 & 67.3 & 83.6 & 74.3 & 88.6 & 83.2 \\
\midrule                                                                                                          
\multicolumn{10}{l}{{Final Score (36 tasks)}}                                                                     \\
\rowcolor[HTML]{FFF2CC} \textit{All IND Avg.}      & 37.1 & 32.3 & 25.3 & 67.5 & 55.8 & 68.0 & 62.3 & 72.3 & 74.7 \\
\rowcolor[HTML]{FFF2CC} \textit{All OOD Avg.}      & 38.7 & 38.0 & 25.1 & 57.1 & 60.9 & 59.1 & 55.5 & 61.5 & 65.9 \\
\rowcolor[HTML]{D9E1F2} \textit{All Tasks Avg.}    & 37.8 & 34.8 & 25.2 & 62.9 & 58.0 & 64.1 & 59.3 & 67.5 & 70.8 \\

\bottomrule
\end{tabular}
}
\label{tab:detail_mmeb}
\end{table*}